\documentclass{article}

\usepackage{arxiv}

\usepackage[utf8]{inputenc} 
\usepackage[T1]{fontenc}    
\usepackage{hyperref}       
\usepackage{url}            
\usepackage{booktabs}       
\usepackage{amsfonts}       
\usepackage{nicefrac}       
\usepackage{microtype}      
\usepackage{lipsum}		
\usepackage{graphicx}
\usepackage{natbib}
\usepackage{doi}

\usepackage{amsmath}
\usepackage{amssymb}
\usepackage{mathtools}
\usepackage{amsthm}

\usepackage{multirow}
\usepackage{stfloats}

\RequirePackage{algorithm}
\RequirePackage{algorithmic}

\usepackage{microtype}
\usepackage{graphicx}
\usepackage{subfigure}
\usepackage{booktabs} 

\usepackage{wrapfig}
\usepackage{caption}
\usepackage[capitalise]{cleveref}

\theoremstyle{plain}

\theoremstyle{definition}

\theoremstyle{remark}

\usepackage{setspace}

\title{Does Deep Active Learning Work in the Wild?}

\author{%
  Simiao Ren\\
  Meta\\
  Menlo Park, CA 94025\\
  \texttt{benren@meta.com} \\
  \And
  Saad Lahrichi \\
  University of Missouri\\
  Columbia, MO 65203 \\
  \texttt{saad.lahrichi@missouri.edu} \\
  \And
  Yang Deng, Willie J. Padilla, Leslie Collins \\
  Duke University\\
  Durham, NC 27705 \\
  \texttt{[first].[last]@duke.edu} \\
  \And
  Jordan M. Malof \\
  University of Missouri\\
  Columbia, MO 65203 \\
  \texttt{jmdrp@missouri.edu} \\
}

\date{}

\hypersetup{
pdftitle={Does Deep Active Learning Work in the Wild?},
pdfsubject={cs, ml},
pdfauthor={Simiao Ren, Saad Lahrichi, Yang Deng, Willie J. Padilla, Leslie Collins, Jordan M. Malof },
pdfkeywords={deep learning, active learning, robustness, sampling, diversity},
}

\begin{document}
\maketitle

\begin{abstract}
Deep active learning (DAL) methods have shown significant improvements in sample efficiency compared to simple random sampling. While these studies are valuable, they nearly always assume that optimal DAL hyperparameter (HP) settings are known in advance, or optimize the HPs through repeating DAL several times with different HP settings. Here, we argue that in real-world settings, or \textit{in the wild}, there is significant uncertainty regarding good HPs, and their optimization contradicts the premise of using DAL (i.e., we require labeling efficiency).  In this study, we evaluate the performance of eleven modern DAL methods on eight benchmark problems as we vary a key HP shared by all methods: the pool ratio.  Despite adjusting only one HP, our results indicate that eight of the eleven DAL methods sometimes underperform relative to simple random sampling and some frequently perform worse. Only three methods always outperform random sampling (albeit narrowly), and we find that these methods all utilize diversity to select samples - a relatively simple criterion.  Our findings reveal the limitations of existing DAL methods when deployed \textit{in the wild}, and present this as an important new open problem in the field. 
\end{abstract}


\section{Introduction}
\label{sec:introduction}

In this work, we focus on the application of active learning to deep neural networks (DNNs), sometimes referred to as Deep Active Learning (DAL) \cite{roy2018deep}. Broadly speaking, the premise of DAL is that some training instances will yield superior performance compared to others. Therefore, we can improve the training sample efficiency of DNNs by selecting the best training instances. A large number of methods have been investigated in recent years for DAL \cite{settles2009active, ren2021survey, holzmuller2023framework}, often reporting significant improvements in sample efficiency compared to simpler strategies, such as random sampling \cite{tsymbalov2018dropout, kading2018active, kee2018query}. While these studies provide valuable insights, they nearly always assume good DAL hyperparameter (HP) settings are known in advance, or alternatively, they optimize the HPs (e.g., by repeating DAL several times with different HP settings).  To our knowledge however, there is little evidence that one can assume good hyperparameters are known in advance for novel problems (see \cref{sec:ProblemSetting}, where we find HP settings in the literature vary widely across problems).  Moreover, running a DAL method multiple times in search of good HP settings may result in significant label inefficiency, even when compared to random sampling. Therefore, in real-world settings where DAL is applied to a novel problem, or \textit{in the wild} as we term it here, the best DAL HPs are not generally known in advance, and it is unclear whether DAL still offers advantages (e.g., compared to random sampling) when accounting for HP uncertainty. If DAL models do not reliably outperform simple random sampling in the presence of HP uncertainty, it greatly undermines their value, and the likelihood that they will be adopted.  Despite the significance of this problem, it has received little attention in the literature.

\begin{figure*} 
    \begin{center}
    \centerline{\includegraphics[width=\linewidth]{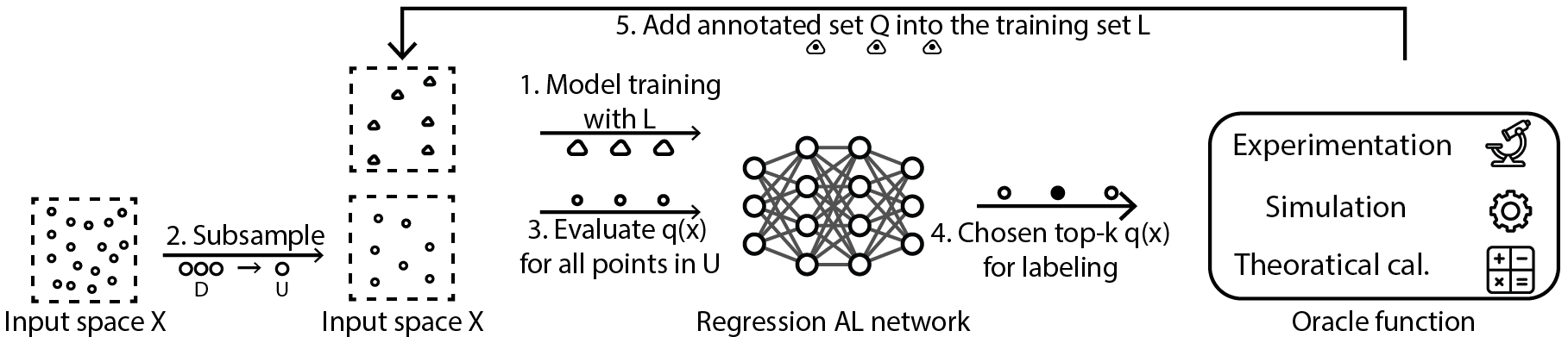}}
    \caption{Schematic diagram for pool-based DAL procedure. In the input space X, the triangles represent labeled data ($L$), and the circles represent unlabeled data ($D$ for the full set of unlabeled data, and$U$ for subsampled unlabeled pool). At each step, after the model is trained using the existing training set $L$, a subset of unlabeled data $U$ is sampled and evaluated by the AL criteria q(x). Then, the top-k points according to q(x) are labeled by the oracle function.}
    \label{img:pool_based}
    \end{center}
\end{figure*}

\paragraph{Contributions}
\label{subsec:contributions_of_this_work}

In this work, we perform the first systematic evaluation of DAL \textit{in the wild}.  We focus our investigation on DAL for regression, where to our knowledge, most applicable DAL methods are pool-based, and therefore they share an important HP: the pool ratio, $\gamma$ (see \cref{sec:ProblemSetting}). Using this property of regression problems, we evaluate a large number of DAL models as we vary a single HP, their $\gamma$ setting, thereby providing a \textit{distribution} of performance that one can expect in real-world settings (i.e., in the wild), where the best setting for $\gamma$ is uncertain. We note that most DAL models have several (often unique) HPs that exhibit uncertainty, and each can contribute to performance variability of DAL methods in the wild.  However, examining variability with respect to all of these HPs would require a lengthy exposition, and would be computationally costly. Therefore we focus on $\gamma$, which mitigates the aforementioned challenges, while still providing sufficient empirical evidence to support our main conclusions.

To support our investigation, we assembled eight scientific computing regression problems to examine the performance of DAL methods in this setting; to our knowledge, this is the first such benchmark of its kind. We then identified past and recent DAL methods that are suitable for regression, totaling eleven methods

To support our study, we identified eleven DAL methods that are suitable for regression. We then examined the performance of these DAL methods on each of eight benchmark problems, compared to simple random sampling, as we vary their $\gamma$ settings. Our results indicate that their performance varies significantly with respect to $\gamma$, and that the best HP varies for different DAL/dataset with no single $\gamma$ value working best across all settings, confirming our hypothesis that there is significant uncertainty regarding the best HP setting for novel problems.  We also find that most of the DAL methods sometimes underperform simple random sampling and some frequently perform much worse:

\begin{itemize}
    
    \item We compile a large benchmark of eleven state-of-the-art DAL methods across eight datasets. For some of our DAL methods, we are the first to adapt them to regression. Upon publication, we will publish the datasets and code to facilitate reproducibility. 
    
    \item Using our benchmark, we perform the first analysis of DAL performance \textit{in the wild}. Using $\gamma$ as an example, we systematically demonstrate the rarely-discussed problem that most DAL models are often outperformed by simple random sampling when we account for HP uncertainty.        
     
    \item We analyze the factors that contribute to the robustness of DAL in the wild, with respect to $\gamma$.    
\end{itemize}

\section{Related works}\label{sec:related_work}

\paragraph{Active learning benchmarks} The majority of existing AL benchmarks are for classification tasks, rather than regression \cite{jose2024regression}, and many AL methods for classification cannot be applied to regression. Some existing studies include \cite{zhan2021comparative}, which benchmarked AL using a Support Vector Machine (SVM) with 17 AL methods on 35 datasets. \cite{yang2018benchmark} benchmarked logistic regression with 12 AL methods and 44 datasets. \cite{meduri2020comprehensive} benchmarked specific entity matching application (classification) of AL with 3 AL methods on 12 datasets, with 3 different types of classifiers (DNN, SVM, and Tree-based). \cite{trittenbach2021overview} benchmarked an AL application in outlier detection on 20 datasets and discussed the limitation of simple metrics extensively. \cite{hu2021towards} benchmarked 5 classification tasks (including both image and text) using DNN. \cite{beck2021effective} benchmarked multiple facets of DAL on 5 image classification tasks. For the regression AL benchmark, \cite{o2017model} benchmarked 5 AL methods and 7 UCI \footnote{University of California Irvine Machine Learning Repository} datasets, but they only employed linear models. \cite{wu2019active} compared 5 AL methods on 12 UCI regression datasets, also using linear regression models. Our work is fundamentally different from both, as we use DNNs as our regressors, and we employ several recently-published problems that also involved DNN regressors, making them especially relevant for DAL study. The recent study by \cite{holzmuller2023framework} is the only work that is similar to ours, in which the authors benchmarked 8 pool-based DAL methods for regression on 15 datasets.  The primary focus of their work was to propose a novel DAL regression framework, termed LCMD; meanwhile the focus of our work is to investigate DAL in the wild. Consequently, \cite{holzmuller2023framework} presents different performance metrics and conclusions compared to our study.     

\paragraph{Active learning for regression problems} Regression problems have received (relatively) little attention compared to classification \cite{ren2021survey, guyon2011results}. For the limited AL literature dedicated to regression tasks, Expected Model Change (EMC) \cite{settles2008curious, cai2013maximizing} was explored, where an ensemble of models was used to estimate the true label of a new query point using both linear regression and tree-based regressors. Gaussian processes were also used with a natural variance estimate on unlabeled points in a similar paradigm \cite{kading2018active}. \cite{smith2018less} used Query By Committee (QBC), which trains multiple networks and finds the most disagreeing unlabeled points of the committee of models trained. \cite{tsymbalov2018dropout} used the Monte Carlo drop-out under a Bayesian setting, also aiming for maximally disagreed points. \cite{yu2010passive} found $x$-space-only methods outperforming y-space methods in robustness. \cite{yoo2019learning} proposed an uncertainty-based mechanism that learns to predict the loss using an auxiliary model that can be used on regression tasks. \cite{ranganathan2020deep} and \cite{kading2016active} used Expected Model Output Change (EMOC) with Convolutional Neural Network (CNN) on image regression tasks with different assumptions. We included all these methods that used deep learning in our benchmark.

\paragraph{DAL in the wild} To our knowledge, all empirical studies of pool-based DAL methods assume that an effective pool ratio hyperparameter, $\gamma$, is known apriori. While the majority of works assumed the original training set as the fixed, unlabeled pool, \cite{yoo2019learning} limited their method to a subset of 10k instances instead of the full unlabeled set and \cite{beluch2018power} used subsampling to create the pool $U$ (and hence $\gamma$). In real-world settings - in the wild - we are not aware of any method to set $\gamma$ a priori, and there has been no study of DAL methods under this setting. Therefore, we believe ours is the first such study.

\section{Problem Setting}  \label{sec:ProblemSetting}

In this work, we focus on DAL for regression problems, which comprise a significant portion of DAL problems involving DNNs \cite{jose2024regression}.  As discussed in \cref{sec:introduction}, nearly all DAL methods for regression are pool-based, which is one of the three major paradigms of AL, along with stream-based and query synthesis. \cite{settles2009active}

\paragraph{Formal description} Let $L^i = (X^{i}, Y^{i})$ be the dataset used to train a regression model at the $i^{th}$ iteration of active learning.  We assume access to some oracle, denoted $f : \mathcal{X} \rightarrow \mathcal{Y}$, that can accurately produce the target values, $y \in \mathcal{Y}$ associated with input values $x \in \mathcal{X}$.  Since we focus on DAL, we assume a DNN as our regression model, denoted $\hat{f}$.  We assume that some relatively small number of $N_{0}$ labeled training instances are available to initially train $\hat{f}$, denoted $L^0$.  In each iteration of DAL, we must choose $k$ query instances to be labeled by the oracle, yielding a set of labeled instances, denoted $Q$, that is added to the training dataset. Our goal is then to choose $Q$ that maximizes the performance of the DNN-based regression models over  unseen test data at each iteration of active learning.  

\paragraph{Pool-based Deep Active Learning} General pool-based DAL methods assume that we have some pool $U$ of $N_{U}$ unlabeled instances from which we can choose the $k$ instances to label. The set $U$ is sampled from a larger and potentially-infinite set, denoted $D$, and $N_{U}$ is a HP chosen by the DAL user.  We note that in some DAL applications, such as computer vision, it is conventional to utilize all available unlabeled data for $U$, and the pool size is not often explicitly varied or discussed. However, this convention is equivalent to setting $U=D$, and thereby implicitly setting the $N_{U}$ HP. Most pool-based methods rely upon some acquisition function $q: \mathcal{X} \rightarrow \mathbb{R}$ to assign some scalar value to each $x \in U$ indicating its "informativeness", or utility for training $\hat{f}$. In each iteration of active learning, $q$ is used to evaluate all instances in $U$, and the top $k$ are chosen to be labeled and included in $L$. 

\begin{figure}[h]
    \begin{center}
    \centerline{\includegraphics[width=\linewidth]{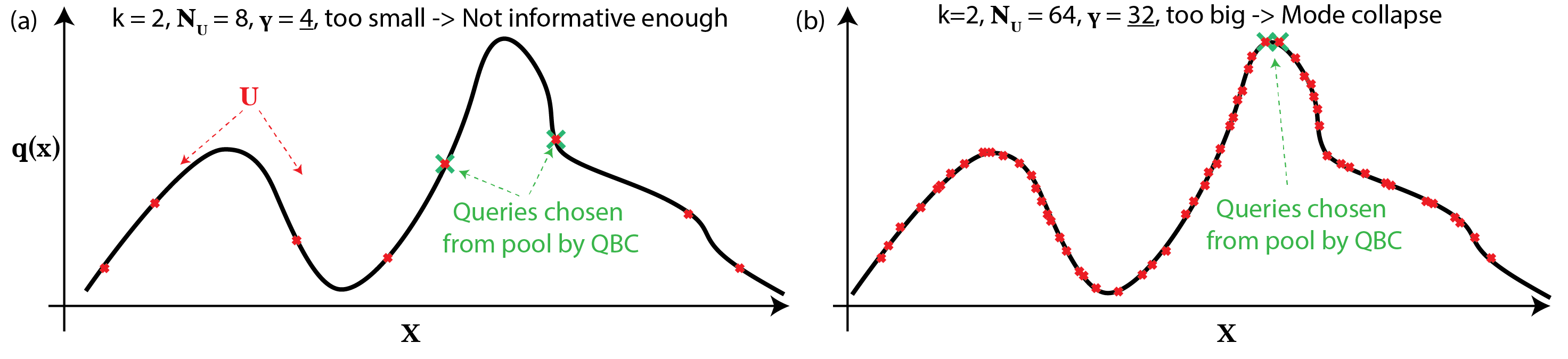}}
    \caption{Pool-based DAL for uncertainty-based mechanism. $q(x)$ is the acquisition metric. (a, b) are two scenarios of the pool ratio ($\gamma$) being too small (4 in a) or too large (32 in b) in $k$ (step size) of 2. }
    \label{img:pool_ratio_schematic}
    \end{center}
\end{figure}

\paragraph{The pool ratio hyperparameter, $\gamma$}
We define the \textit{pool ratio} as $\gamma = N_{U}/k$.  By definition, $N_{U}$ and $k$ are hyperparameters of pool-based problems, and therefore $\gamma$ also is.  While one could, in principle, vary $N_{U}$ and $k$ independently, this is not often done in practice.  Typically $k$ is set as small as possible, limited by computational resources.  This leaves $N_{U}$ as the major free hyperparameter; however, prior research has found that its impact depends strongly on its size relative to $k$ \cite{kee2018query, tsymbalov2018dropout, kading2018active}, encoded in $\gamma$.  Given a fixed value of $k$, increasing $N_{U}$ can lead to the discovery of points with higher values of $q(x)$ due to denser sampling of the input space. However, a larger $N_{U}$ also increases the similarity of the points, which provides redundant information to the model - a problem referred to as mode collapse \cite{burbidge2007active, ren2021survey, kee2018query}. In the limit as $N_{U} \rightarrow \infty$, all of the $k$ selected query points will be located near the same $x \in \mathcal{X}$ that has the highest value of $q(x)$. This tradeoff is illustrated in \cref{img:pool_ratio_schematic} for a simple problem, and has also been noted in \cite{cacciarelli2024active}. 

In most real-world settings, there is a substantial quantity of unlabeled data (often infinite), and the user has the freedom (or burden) of choosing a suitable $\gamma$ setting for their problem by varying the size of $U$.  Crucially, and as we show in our experiments, choosing a sub-optimal $\gamma$ value can result in poorer performance than naive random sampling. This is not necessarily a problem if either (i) one $\gamma$ setting works across most problems or, alternatively, (ii) $\gamma$ can be optimized on new problems without using labels.  To the best of our knowledge, there is no method for optimizing $\gamma$ on a new problem without running multiple trials of AL to find the best one (i.e., collecting labels), defeating the purpose of AL in real-world settings. Furthermore, the value of $\gamma$ varies widely across the literature, suggesting that suitable settings for $\gamma$ indeed vary across problems (see supplement for a list).

\section{Benchmark Regression Problems} 
\label{sec:benchmark_problems}
\begin{table}[h]
    \caption{Benchmark datasets dimensionality and oracle functions. $Dim_{x, y}$ are the dimensionality of $x$ and y. Note that ODE solutions are implemented in the form of analytical functions as well.}
    \label{tbl:benchmark_dataset}
    \begin{center}
        \begin{small}
        \begin{sc}
            \begin{tabular}{lcccccccc}
            \toprule
            Dataset & Sine & Robo & Stack & ADM & Foil & Hydr & Bess & Damp \\
            \midrule
            $Dim_{x}$ & 1 & 4 & 5 & 14 & 5 & 6 & 2 & 3 \\ 
            $Dim_{y}$ & 1 & 2 & 201 & 2000 & 1 & 1 & 1 & 100\\
            Oracle & \multicolumn{2}{c}{Analytical} &  Numerical simulator &  DNN & \multicolumn{2}{c}{Random Forest} & \multicolumn{2}{c}{ODE solution} \\
            \bottomrule
            \end{tabular}
        \end{sc}
        \end{small}
    \end{center}
\end{table}
To compose our benchmarks, we focused primarily upon problems in scientific computing, which is an important emerging problem setting \cite{subramanian2024towards, takamoto2022pdebench, majid2024mixture}. We propose eight regression problems to include in our benchmark set: two simple toy problems (SINE, ROBO), four contemporary problems from publications in diverse fields of science and engineering (STACK, ADM, FOIL, HYDR) and two problems solving ordinary differential equations (also prevalent in engineering). Summary details of our benchmark problems can be found in \cref{tbl:benchmark_dataset} and \cref{tbl:oracle_details}.

We utilized four major selection criteria, beyond choosing scientific computing problems: (i) diversity: we sought to include a set of problems that span different disciplines (aero and fluid-dynamics, materials science), and problems that require physical experiments (e.g., FOIL, HYDRO) versus simulators (e.g., ADM); (ii) availability of labeled data: the problems we chose (unlike many high dimension ones) all had sufficiently large amount of labeled data, allowing us to easily study the impact of different pool ratios; (iii) dimensionality: we sought problems with relatively low dimensionality because they mitigate computational costs allowing for more extensive experimentation, while still being representative of many contemporary scientific computing problems (e.g., labeling can be highly expensive, severely limiting total labeled data, and making even low-dimensional problems challenging); (iv) difficulty: the problems in our dataset are also “difficult” in the sense that the accuracy of the learners (i.e., the DNN regressors) can vary significantly depending upon which data are labeled, making it possible to distinguish between more/less effective AL approaches.  Although this is not the only notion of “difficulty” that may be relevant for selecting benchmark problems, we believe this is the most important one, and has been used in recent DAL studies \cite{holzmuller2023framework}. We now describe our benchmark problems: 

\begin{table}[h]
    \caption{Details of used oracles. Along with details in our repository (to be made public after publication), we provide information on the oracles' source publication, type of ML model, source of training data, quantity of data, and error level. For all datasets, we adopt an 80-20 train-test split.}
    \label{tbl:oracle_details}
    \begin{center}
        \begin{small}
        \begin{sc}
         \resizebox{\textwidth}{!}{%
        \begin{tabular}{lcccc}
            \toprule
            Dataset & Type of ML Model & Source of Data & Quantity of Data & Test MSE \\
            \midrule
            ADM \cite{deng2021neural}   & Ensemble of DNNs    & Numerical simulator  & 160K samples  & 6.00E-05 \\
            FOIL \cite{Dua:2019}  & Random Forest & Real-world experiments & 1503 samples & 8.63e-03 \\
            HYDRO \cite{Dua:2019} & Random Forest & Real-world experiments & 302 samples  & 3.97e-02 \\
            \bottomrule
            \end{tabular}
            }
        \end{sc}
        \end{small}
    \end{center}
\end{table}
\textbf{1D sine wave (SINE)} A noiseless 1-dimensional sinusoid with smoothly-varying frequency.  \textbf{2D robotic arm (ROBO)} \cite{ren2020benchmarking} The goal is to predict the 2-D spatial location of the endpoint of a robotic arm based on its three joint angles.  \textbf{Stacked material (STACK)} \cite{Chen2019} The goal is to predict the 201-D reflection spectrum of a material based on the thickness of its five layers.  \textbf{Artificial Dielectric Material (ADM)} \cite{deng2021neural} The goal is to predict the 2000-D reflection spectrum of a material based on its 14-D geometric structure. Full wave electromagnetic simulations were utilized in \cite{deng2021benchmarking} to label data in the original work, requiring 1-2 minutes per input point. \textbf{NASA Airfoil (FOIL)} \cite{Dua:2019} The goal is to predict the sound pressure of an airfoil based on the structural properties of the foil, such as its angle of attack and chord length.  This problem was published by NASA \cite{brooks1989airfoil} and the instance labels were obtained from a series of real-world aerodynamic tests in an anechoic wind tunnel. It has been used in other AL literature \cite{wu2018pool, liu2020unsupervised, jose2024regression}. \textbf{Hydrodynamics (HYDR)} \cite{Dua:2019} The goal is to predict the residual resistance of a yacht hull in water based on its shape.  This problem was published by the Technical University of Delft, and the instance labels were obtained by real-world experiments using a model yacht hull in the water.  It is also referred to as the "Yacht" dataset in some AL literature \cite{wu2019active, cai2013maximizing, jose2024regression}. \textbf{Bessel function (BESS)} The goal is to predict the value of the solution to Bessel's differential equation, a second-order ordinary differential equation that is common in many engineering problems. The inputs are the function order $\alpha$ and input position $x$. The order $\alpha$ is limited to non-negative integers below 10. \textbf{Damping Oscillator (DAMP)} The goal is to predict the full-swing trajectory of a damped oscillator in the first 100 time steps, of the solution to a second-order ordinary differential equation. The input is the magnitude, damping coefficient, and frequency of the oscillation.

\begin{table*}[t!]
    \caption{List of benchmarked methods. $L$ is the labeled set, $Q$ is the already selected query points with $dist$ being L2 distance, $\hat{f}(x)$ is model estimate of $x$, $f(x)$ is oracle label of $x$, $\mu(x)$ is the average of ensemble model output, N is number of models in the ensemble, $N_k$ is the k-nearest-neighbors, $sim$ is cosine similarity, $\phi$ is current model parameter, $\phi'$ is the updated parameter, $\mathcal{L}(\phi;(x',y'))$ is the loss of model with parameter $\phi$ on new labeled data $(x', y')$, $f_{loss}(x)$ is the auxiliary model that predicts the relative loss}
    \label{tbl:benchmark_method}
    \begin{center}
        \begin{small}
        \begin{sc}
            \begin{tabular}{cc}
            \toprule
            Method  & Acquisition function (q) \\
            \midrule
            Core-set (GSx)  &
            \multirow{2}{*}{$ \displaystyle \min_{x\in \mathcal{L} \cup \mathcal{Q}} dist(x^*, x) $}\\
            \cite{sener2017active} &\\
            \hline
            Greedy sampling in y (GSy)  &  
            \multirow{2}{*}{$\displaystyle  \min_{y\in \mathcal{L} \cup \mathcal{Q}} dist(\hat{f}(x^*), y)$}\\
            \cite{wu2019active} &\\
            \hline
            Improved greedy sampling (GSxy)   &  
            \multirow{2}{*}{$\displaystyle  \min_{(x,y)\in \mathcal{L} \cup \mathcal{Q}} dist(x^*, x)*dist(\hat{f}(x^*), y)$}\\
             \cite{wu2019active}&\\
            \hline
            Query by committee (QBC)   &  
            \multirow{3}{*}{$\displaystyle  \frac{1}{N}\sum^N_{n=1}(\hat{f}_n(x^*)-\mu(x^*))^2$} \\
            \multirow{2}{*}{\cite{kee2018query}} &\\
            &\\
            \hline
            QBC w/ diversity  &  
            $\displaystyle  q_{QBC}(x^*) + q_{div}(x^*) $ \\
            (QBCDiv) \cite{kee2018query} & $\displaystyle (q_{div}(x^*) =  q_{GSx}(x^*) )$ \\
            \hline
            QBC w/ diversity \& density &  
            $\displaystyle  q_{QBC}(x^*) + q_{div}(x^*) + q_{den}(x^*)$  \\
              (QBCDivDen) \cite{kee2018query}&  $\displaystyle  (q_{den}(x^*) = \dfrac{1}{k} \sum_{x\in N_k(x^*)} sim(x^*, x) )$ \\
            \hline
            bayesian by disagreement (BALD)   & \multirow{2}{*}{$\displaystyle  q_{QBC}(x^*)$ with dropout} \\
            \cite{tsymbalov2018dropout} &\\
            \hline
            Expected model output  change  & 
            $\displaystyle \mathbb{E}_{y'|x'} \mathbb{E}_{x} || \hat{f}(x^*; \phi') - \hat{f}(x^*; \phi)||_1 $ \\
             (EMOC) \cite{ranganathan2020deep}  & $\displaystyle \approx \mathbb{E}_{x} || \nabla_{\phi} \hat{f}(x; \phi) * \nabla_{\phi} \mathcal{L}(\phi; (x^{*'}, y'))||_1$ \\
            \hline
            Learning Loss   \cite{yoo2019learning} & $\displaystyle f_{loss}(x^*)$  \\
            \hline
            Cluster-Variance  & \multirow{2}{*}{$q_{QBC}(x)^*$ in clusters}\\
            \cite{citovsky2021batch} & \\
            \hline
            Density-Aware  Core-Set & \multirow{2}{*}{$q_{GSx}(x^*) + q_{den}(x^*)$}\\ 
             (DACS)\cite{kim2022defense}  & \\
            \bottomrule
            \end{tabular}
        \end{sc}
        \end{small}
    \end{center}
    \vskip -0.1 in
\end{table*}

From the literature, we found eleven AL methods that are (i) applicable to regression problems, (ii) with DNN-based regressors, making them suitable for benchmark regression problems. Due to space constraints, we list each method in \cref{tbl:benchmark_method} along with key details, and refer readers to the supplement for full details.  Some of the methods have unique HPs that must be set by the user.  In these cases, we adopt HP settings suggested by the methods' authors, shown in \cref{tbl:benchmark_method}.  Upon publication, we will publish code for all of these methods to support future benchmarking.

\section{Benchmark Experiment Design} \label{sec:exp_design}

In our experiments, we compare eleven state-of-the-art DAL methods on eight scientific computing problems. We evaluate the performance of our DAL methods as a function of $\gamma$ on each of our benchmark problems, with $\gamma \in [2,4,8,16,32,64]$ (i.e., at each step we sample our $U$ with $k*\gamma$ points). Following convention \cite{kee2018query,tsymbalov2018dropout}, we assume a small training dataset is available at the outset of active learning, $T^{0}$, which has $N_{0} = 80$ randomly sampled training instances. We then run each DAL model to $T^{50}$ AL steps, each step identifying $k=40$ points to be labeled from a fresh, randomly generated pool of size $k*\gamma$. For each benchmark problem, we assume an appropriate neural network architecture is known apriori. Each experiment (i.e., the combination of dataset, DAL model, and $\gamma$ value) is run 5 times to account for randomness. The MSE is calculated over a set of 4000 test points that are uniformly sampled within the $x$-space boundary. To reduce unnecessary noise related to our core hypothesis, we use the same (randomly sampled) unlabeled pools across different DAL methods.

We must train a regression model for each combination of problem and DAL method. Because some DAL methods require an ensemble model (e.g., QBC), we use an ensemble of 10 DNNs as the regressor for all of our DAL algorithms (except for the ADM problem, which is set to 5 due to the GPU RAM limit). More details on the models used and training procedures can be found in the supplement.  Following convention \cite{kading2018active, wu2018pool, o2017model}, we summarize our DAL performance by the area under curve (AUC) of the error plot.  We report the full MSE vs \# labeled point plots in the supplement. For the AUC calculation, we use 'sklearn.metrics.auc' \cite{scikit-learn} then further normalize by such AUC of random sampling method for easier visualization. All reported results are given in the unit of normalized AUC of MSE ($nAUC_{MSE})$.

\section{Experimental Results} \label{sec:result}

\begin{figure*}[t!]
    \begin{center}
    \vskip -0.1in
    \centerline{\includegraphics[width=\linewidth]{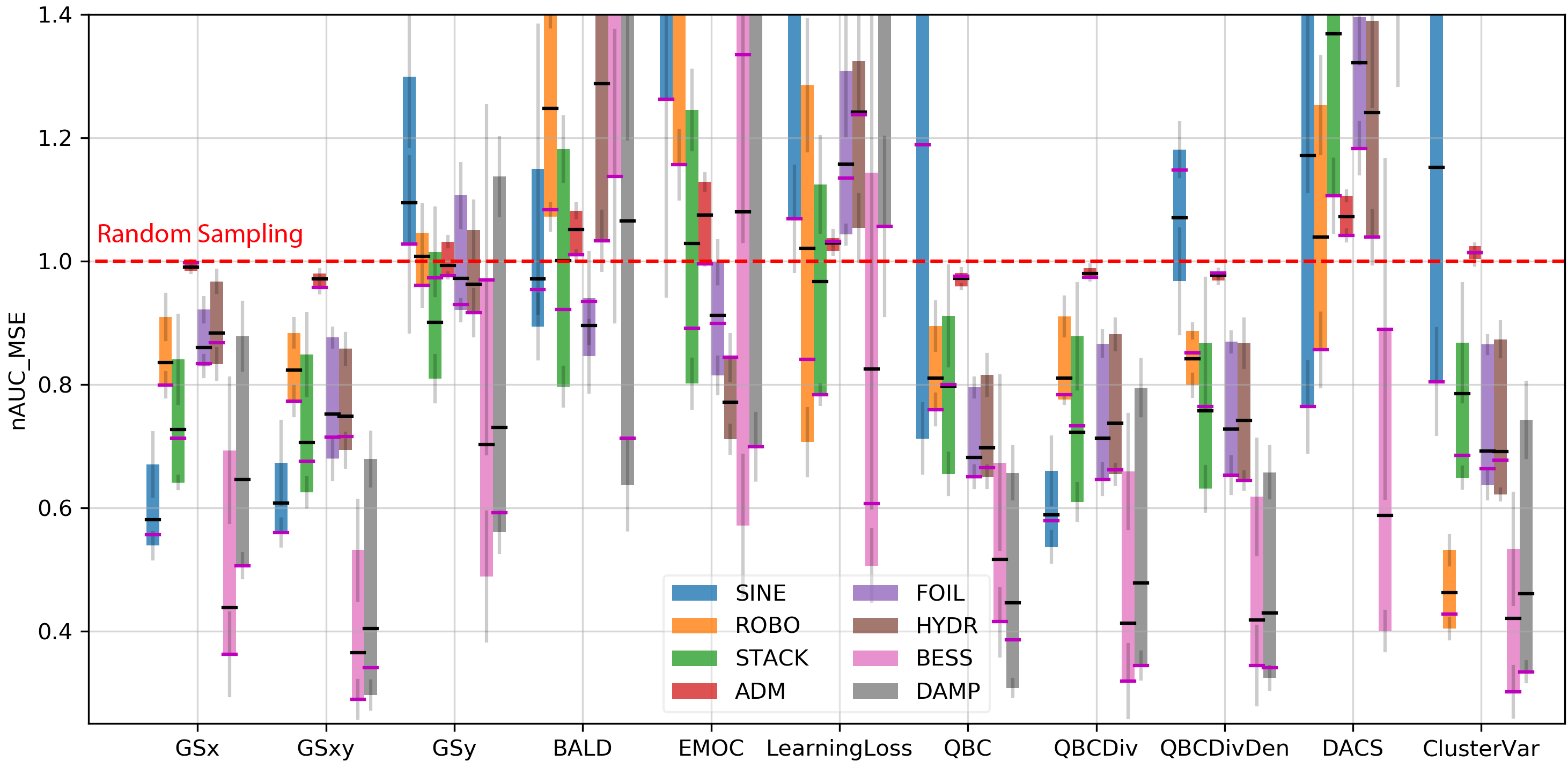}}   
    \caption{Performance of each DAL method (x-axis) in terms of $nAUC_{MSE}$ (y-axis).  For each DAL method, we report a bar indicating the \textit{range} of $nAUC_{MSE}$ values obtained as we vary the pool ratio, $\gamma \in [2,4,...,64]$; for a given DAL method, we report one bar for each of the eight benchmark problems, indicated by a unique color in the legend.  Each bar is bisected by a solid black and magenta line, respectively.  The black line represents the average $nAUC_{MSE}$ value across all settings of $\gamma$.  The magenta line represents the performance using $\gamma_{prior}$ (see \cref{sec:result} for details). The dashed red line at $nAUC_{MSE}=1$ corresponds to the performance obtained using random sampling. Note that some vertical bars are intentionally clipped at the top to improve the visualization overall.}
        \label{img:main_perf}
    \end{center}
    \vskip -0.3in
\end{figure*}

The performance of all eleven DAL methods on all eight benchmark datasets is summarized in \cref{img:main_perf}. The y-axis is the normalized $nAUC_{MSE}$, the x-axis is the DAL methods of interest, and the color code represents the different benchmark datasets.  The horizontal red dashed line represents the performance of random sampling, which by definition is equal to one (see \cref{sec:exp_design}). Further details about \cref{img:main_perf} are provided in its caption. We next discuss the results, with a focus on findings that are most relevant to DAL in the wild.

The results in \cref{img:main_perf} indicate that \textit{all} of our benchmark DAL methods are sensitive to their setting of $\gamma$ - a central hypothesis of this work.  As indicated by the vertical bars in \cref{img:main_perf}, the $nAUC_{MSE}$ obtained by each DAL method varies substantially with respect to $\gamma$.  For most of the DAL methods, there exist settings of $\gamma$ (often many) that cause them to perform worse than random sampling. This has significant implications for DAL in the wild since, to our knowledge, there is no general method for estimating a good $\gamma$ setting prior to collecting large quantities of labeled data (e.g., to run trials of DAL with different $\gamma$ settings), and DAL methods may perform worse, and unreliably, when accounting for the uncertainty of $\gamma$.  

\begin{figure}
    \begin{center}
    \centerline{\includegraphics[width=0.65\textwidth]{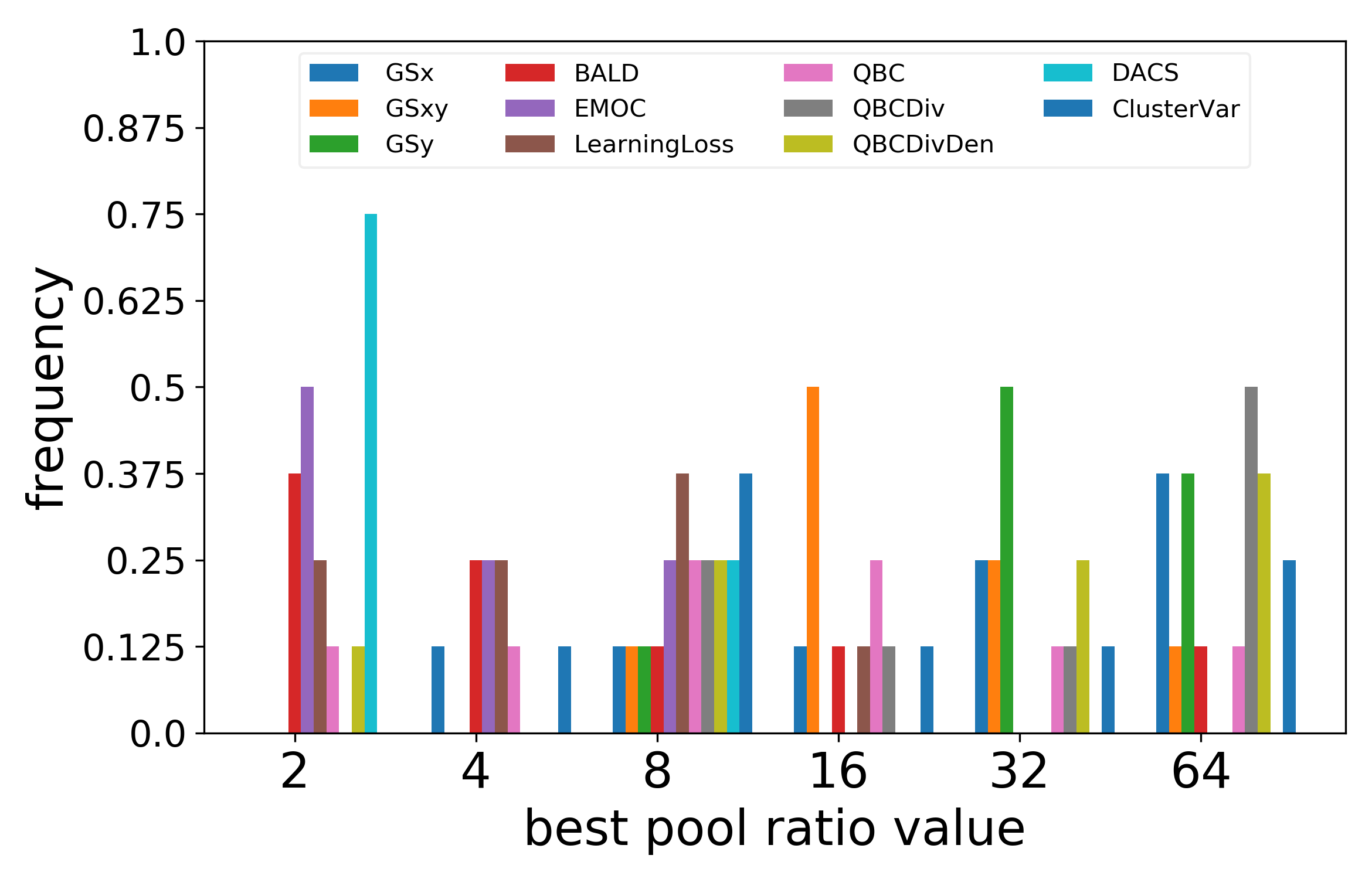}}
    \caption{Frequency histogram of the best pool ratio values found in each DAL. For a given DAL method, this figure shows the frequency (\% out of 8) that a particular pool ratio (x-axis) performs the best in terms of average $nAUC_{MSE}$ metric. }
    \label{img:best_pr_hist}
    \end{center}
\end{figure}
\subsection{DALs are sensitive to their pool ratio, $\gamma$}
\label{sec:results_dals_are_sensitive_to_pool_ratio}

The sensitivity of DAL regression models to $\gamma$ may be less significant if there exist $\gamma$ settings that tend to perform well across most problems (for a given DAL method). \cref{img:best_pr_hist} presents a histogram of the best-performing $\gamma$ settings for each DAL method.  The results indicate that for each method there is no setting of $\gamma$ that performs best across all problems.  This corroborates our observations from the literature where we found a wide range of $\gamma$ settings used across studies. However, we do see that some methods tend to have similar $\gamma$ settings across all problems.  For example, DACS has its best performance near $\gamma=2$, although DACS performs poorly overall. GSxy, however, is one of the best-performing methods overall, and its best-performing settings cluster around $\gamma = 16$.  Given this observation, we investigate how well we can perform if we use historical results for a given method to choose a $\gamma$ value for future problems.  We emulate this scenario by evaluating the performance of each DAL method when adopting the best single $\gamma$ setting from \cref{img:best_pr_hist} (i.e., the setting that wins across the most benchmarks), which we term $\gamma_{prior}$, and then apply it across all benchmarks. The result of this strategy is given by the magenta line in \cref{img:main_perf}.  In most (but not all) cases, $\gamma_{prior}$ yields lower MSE than the average MSE of all $\gamma$ settings (the black lines).  In some cases, $\gamma_{prior}$ yields substantial overall performance improvements, such as for GSx and GSxy, suggesting that this is a reasonable $\gamma$ selection strategy, although the benefits seem to vary across DAL models.  However, even when using $\gamma_{prior}$, the performance of DAL models still varies greatly, and many models still perform worse than random sampling. Therefore, while $\gamma_{prior}$ may often be beneficial, it does not completely mitigate $\gamma$-uncertainty.  

\subsection{Do any DAL methods outperform random sampling in the wild?}
\label{sec:results_do_dal_outperform_random_sampling}
The results indicate that several DAL methods \textit{tend} to obtain much lower $nAUC_{MSE}$  (i.e., they are better) than random sampling. This includes methods such as GSx, GSxy, GSy, QBC-x (variations of QBC) and ClusterVar.  The results therefore suggest that these methods are beneficial more often than not, compared to random sampling - an important property.  However, as discussed in \cref{sec:results_dals_are_sensitive_to_pool_ratio}, all DAL methods exhibit significant performance variance with respect to $\gamma$, and some of the aforementioned methods still sometimes perform worse than random sampling.  For example, this is the case of QBC, GSy, and QBCDivDen on the SINE problem.  In settings where DAL is useful, the cost of collecting labels tends to be high, and therefore the risk of poor DAL performance (e.g., relative to simple random sampling) may strongly deter its use. Therefore, another important criteria is performance robustness: do any DAL methods consistently perform better than random sampling, in the wild?  Our results indicate that GSx, GSxy, and QBCDiv always perform at least as well as random sampling, and often substantially better, regardless of the problem or $\gamma$ setting.  Note that all three robust DALs (GSx, GSxy, QBCDiv) employ x-space diversity in their loss function, which we discuss further in \cref{sec:results_why_some_methods_are_better}.

\begin{figure*}[t!]
    \begin{center}
    \centerline{\includegraphics[width=\linewidth]{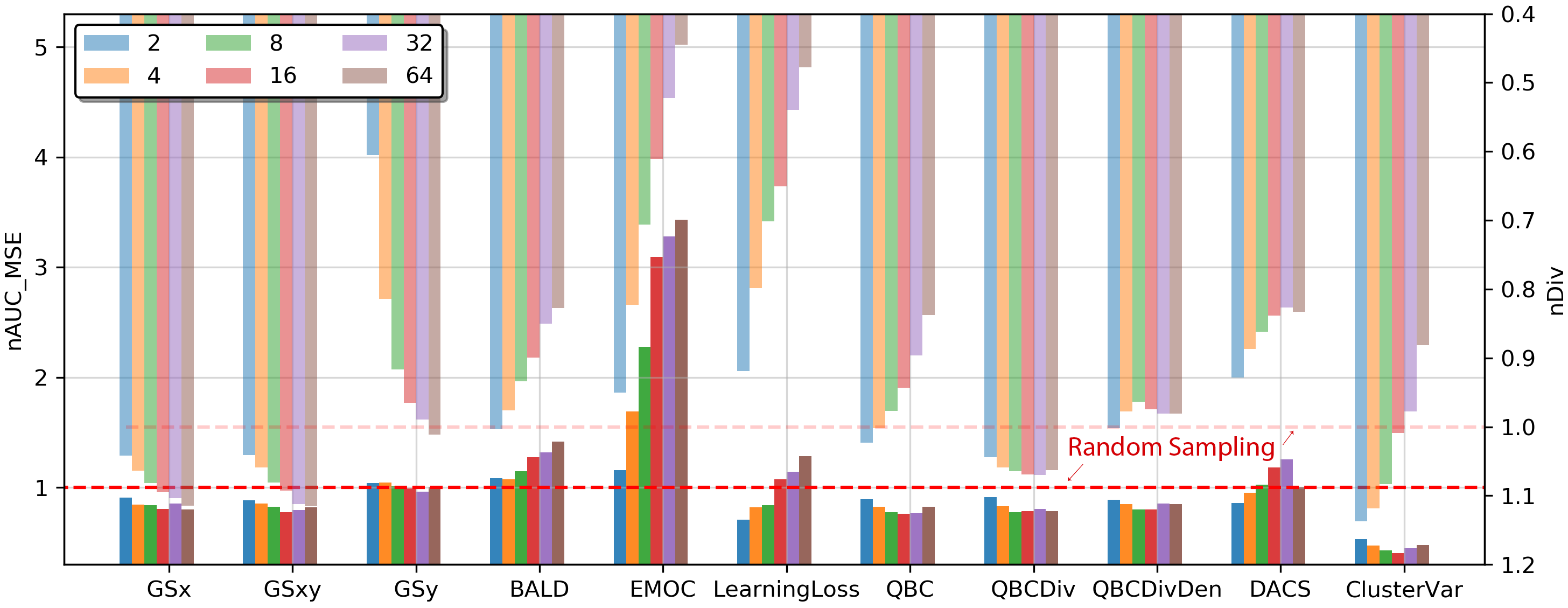}}
    \caption{A representative, combined plot with $nAUC_{MSE}$ performance (bottom, y-axis at left, solid) and collapse metric, nDiv (upper, y-axis at right, more transparent) for each of the eleven DAL models at all pool ratios (color coded) for \textit{robotic arm} dataset (ROBO). Dashed horizontal red lines starting from both $y$ axes represent the random sampling's average $nAUC_{MSE}$ and nDiv at 1. }
    \label{img:mode_collapse}
    \end{center}
\end{figure*}

\subsection{Sample diversity is important for DAL in the wild} \label{sec:results_why_some_methods_are_better}
Our results indicate that the best-performing DAL methods are GSx, GSxy, and QBCDiv. We say these methods are "best" because they are both robust (see \cref{sec:results_do_dal_outperform_random_sampling}), and they also usually yield lower MSEs, than other DAL methods.  These methods share the common property that they encourage training data diversity, as measured by $x$-space distance between points. Interestingly, GSx \textit{only} relies on x-space diversity.  These results suggest that $x$-space diversity is a highly effective DAL acquisition criterion.  Furthermore, and in contrast to other criteria, seeking points that maximize $x$-space diversity does not (by definition) increase the risk of mode collapse. Consequently, increasing $\gamma$ results in greater diversity but without any increased risk of mode collapse (more details in \cref{img:mode_collapse}).  This may be a major reason why GSx, GSxy, and QBCDiv are less sensitive to $\gamma$, and provide much more robust performance in the wild than other DAL methods.  While sampling methods that use diversity have been found to be promising \cite{jose2024regression}, our work provides evidence, for the first time, that sampling based upon diversity \textit{may} be robust to hyperparameter uncertainty (we only examine uncertainty of $\gamma$) while other popular sampling criteria (e.g., estimated model error) seem to be much less reliable in the wild.

To corroborate these findings, we evaluated the $x$-space diversity of each DAL method as a function of $\gamma$. In particular, we calculated the diversity metric as the average nearest neighbor distance
\begin{equation*}
    Div = \dfrac{1}{|T|} \sum_t^T \dfrac{1}{K} \sum_{i}^K \min_{x^* \in \mathcal{Q^T}} dist(x^*, x^i)
\end{equation*}
where $\mathcal{Q}^t$ represents the queried batch at active learning step $t$ and $|T|=50$ is the total number of active learning steps. Note that this metric is similar to, but not a simple average of $q_{GSx}(x)$ as $Div$ only focuses on per batch diversity and does not take the labeled set into consideration. It is also further normalized ($nDiv$) by the value of random sampling for each dataset separately. The lower this metric's value, the more severe the mode collapse issue would be.

The $nDiv$ is plotted in the top half of \cref{img:mode_collapse} using the inverted right y-axis. For the obvious failure cases (BALD, EMOC and Learning Loss) in this particular dataset (their $nAUC_{MSE}$ exceeds 1), a clear trend of mode collapse can be observed in the upper half of the plot (nDiv much lower than 1). Meanwhile, a strong correlation between the pool ratio and the diversity metric can be observed: (i) For GSx and GSxy methods, which seek to maximize diversity, their diversity increases monotonically with larger pool ratio. (ii) For uncertainty-based methods (BALD, EMOC, LearningLoss, QBC, MSE), which seek to maximize query uncertainty, their diversity decreases monotonically with larger pool ratios. (iii) For combined methods like QBCDiv and QBCDivDen, the relationship between pool ratio and diversity shows a weak correlation, consistent with the benefits of having diversity as a selection criterion. (iv) Lastly, we observe that all top-performing methods have high diversity, regardless of $\gamma$, suggesting it is an important condition for effective DAL.

\section{Conclusions} \label{sec:conclusions}
For the first time, we evaluated eleven state-of-the-art DAL methods on eight benchmark datasets for regression \textit{in the wild}, where we assume that the best pool ratio hyperparameter, $\gamma$, is uncertain. We summarize our findings as follows: 
\begin{itemize} 
\item \textit{DAL methods for regression often perform worse than simple random sampling, when evaluated in the wild}.  Using $\gamma$ as an example, we systematically demonstrate the rarely-discussed problem that most DAL models are often outperformed by simple random sampling when we account for HP uncertainty.  
\item \textit{Some DAL methods were relatively robust, and outperformed random sampling robustly in the wild (e.g., GSx, GSxy, QBCDiv)}.  
\item \textit{Insofar as robustness to pool ratio is concerned, our results suggest that DAL approaches utilizing sample diversity tend to be much more robust in the wild than other popular selection criteria.}    
\end{itemize}
\subsection{Limitations} One limitation of this work is that we focused on scientific computing benchmark problems, and problems with relatively low dimensionality. Including higher dimensional problems is an especially important opportunity for future work due to the importance of vision problems in the DAL community, and also because sensitivity to pool ratio has been noted in that setting as well \cite{yoo2019learning, sener2017active}, but not studied systematically.  Another important limitation is that we constrained our evaluation of DAL methods to uncertainty in their pool ratio. Future studies would benefit from evaluating each DAL approach with respect to uncertainty in all of its relevant DAL HPs (i.e., those that require labeled data to be optimized), providing a more comprehensive assessment of modern DAL methods in the wild.

\subsubsection*{Acknowledgments}
Simiao Ren thanks the support of the Duke aiM trainee program, by the NSF grant DGE-2022040.

\bibliography{references}\bibliographystyle{iclr2025_conference}

\newpage

\appendix

\section{Details of the benchmarking methods}

\textbf{Core-set (GSx: Greedy sampling in $x$ space)} \cite{sener2017active}. This approach only relies upon the diversity of points in the input space, $\mathcal{X}$, when selecting new query locations. A greedy selection criterion is used, given by
\begin{equation*}
    q_{GSx}(x^*) = \min_{x\in \mathcal{L} \cup \mathcal{Q}} dist(x^*, x)
\end{equation*}
where $\mathcal{L}$ is the labeled set, $\mathcal{Q}$ is the already selected query points and $dist$ being L2 distance.

\textbf{Greedy sampling in $y$ space (GSy)} \cite{wu2019active}. Similar to GSx which maximizes diversity in the $x$ space in a greedy fashion, GSy maximizes the diversity in the $y$ space in a greedy fashion:
\begin{equation*}
    q_{GSy}(x^*) = \min_{y\in \mathcal{L} \cup \mathcal{Q}} dist(f(x^*), y)
\end{equation*}
where $f(x)$ is the current model prediction of the $x$ and $y$ is the labels in the already labeled training set plus the predicted labels for the points (to be labeled) selected in the current step.

\textbf{Greedy sampling in xy space (GSxy)} \cite{wu2019active}. Named as 'Improved greedy sampling (iGS)' in the original paper \cite{wu2019active}, this approach combines GSx and GSy and uses multiplication of the distance of both $x$ and $y$ space in its acquisition function:
\begin{equation*}
    q_{GSxy}(x^*) = \min_{(x,y)\in \mathcal{L} \cup \mathcal{Q}} dist(x^*, x)*dist(f(x^*), y)
\end{equation*}

\textbf{Query-by-committee (QBC)} \cite{seung1992query} The QBC approach is pure uncertainty sampling if we set $q(x) = q_{QBC}(x)$:
\begin{equation*}
    q_{QBC}(x) = \frac{1}{N}\sum^N_{n=1}(\hat{f}_n(x)-\mu(x))^2
\end{equation*}
Here $\hat{f}_{n}$ denotes the $n^{th}$ model in an ensemble of $N_{ens}$ models (DNNs in our case), and $\mu(x)$ is the mean of the ensemble predictions at $x$. In each iteration of AL these models are trained on all available training data at that iteration. 

\textbf{QBC with diversity (Div-QBC)} \cite{kee2018query}.  This method improves upon QBC by adding a term to $q$ that also encourages the selected query points to be diverse from one another. This method introduces a hyperparameter for the relative weight of the diversity and QBC criteria and we use an equal weighting ($\alpha = 0.5$ \cite{kee2018query}). 
\begin{align*}
    q_{QBCDiv}(x) &= (1-\alpha)* q_{QBC}(x) + \alpha * q_{div}(x)\\
    q_{div}(x^*) &=  q_{GSx}(x^*) 
\end{align*}

\textbf{QBC with diversity and density (DenDiv-QBC)} \cite{kee2018query}.  This method builds upon Div-QBC by adding a term to $q(x)$ that encourages query points to have uniform density. This method introduces two new hyperparameters for the relative weight ($\alpha = \beta = \dfrac{1}{3}$) of the density, diversity, and QBC criteria, and we use an equal weighting as done in the original paper \cite{kee2018query}. 
\begin{align*}
    q_{QBCDivDen}(x) &= (1-\alpha - \beta)* q_{QBC}(x) \\
    & + \alpha * q_{div}(x) + \beta * q_{den}(x)\\
    q_{den}(x^*) &= \dfrac{1}{k} \sum_{x\in N_k(x^*)} sim(x^*, x)
\end{align*}
where $N_k(x^*)$ is the k nearest neighbors of an unlabeled point, $sim(x^*, x)$ is the cosine similarity between points.

\textbf{Bayesian active learning by disagreement (BALD)} \cite{tsymbalov2018dropout}. BALD uses the Monte Carlo dropout technique to produce multiple probabilistic model output to estimate the uncertainty of model output and uses that as the criteria of selection (same as $q_{QBC}(x)$). We used 25 forward passes to estimate the disagreement. 

\textbf{Expected model output change (EMOC)} \cite{kading2018active, ranganathan2020deep}. EMOC is a well-studied AL method for the classification task that strives to maximize the change in the model (output) by labeling points that have the largest gradient. However, as the true label is unknown, some label distribution assumptions must be made. Simple approximations like uniform probability across all labels exist can made for classification but not for regression tasks. \cite{ranganathan2020deep} made an assumption that the label is simply the average of all predicted output in the unlabeled set ($y'(x') = \mathbb{E}_{x \in \mathcal{U}} f(x)$) and we use this implementation for our benchmark of EMOC.
\begin{align*}
    q_{EMOC}(x') &= \mathbb{E}_{y'|x'} \mathbb{E}_{x} || f(x; \phi') - f(x; \phi)||_1\\
    &\approx \mathbb{E}_{x} || \nabla_{\phi} f(x; \phi) * \nabla_{\phi} \mathcal{L}(\phi; (x', y'))||_1
\end{align*}
where $f(x; \phi)$ is the current model output for point $x$ with model parameter $\phi$, $\phi'$ is the updated parameter after training on labeled point x' with label y' and $\mathcal{L}(\phi; (x', y')$ is the loss of the model with current model parameter $\phi$ on new labeled data $(x', y')$.

\textbf{Learning Loss} \cite{yoo2019learning}. Learning Loss is another uncertainty-based AL method that instead of using proxies calculated (like variance), learns the uncertainty directly by adding an auxiliary model to predict the loss of the current point that the regression model would make. The training of the auxiliary model concurs with the main regressor training and it uses a soft, pair-wise ranking loss instead of Mean Squared Error (MSE) loss to account for the fluctuations of the actual loss during the training.
\begin{equation*}
    q_{LL}(x) = f_{loss}(x)
\end{equation*}
where $f_{loss}(x)$ is the output of the loss prediction auxiliary model. In this AL method, there are multiple hyper-parameters (co-training epoch, total auxiliary model size, auxiliary model connections, etc.) added to the AL process, all of which we used the same values in the paper if specified \cite{yoo2019learning}.

\textbf{Density Aware Core-set (DACS)} \cite{kim2022defense}
A diversity-based AL method that not only considers core-set metric but also considers the density and strives to sample low-density regions. The original DACS also encodes the image space into feature space and uses locality-sensitive hashing techniques to accelerate the nearest neighbor calculation and prevent computational bottlenecks. As our scientific computing tasks does not involve high dimensional image as well as having much lower dataset size in general, instead of encoded feature space distance, we used input space distance and locality-sensitive hashing was dropped as we don’t face such computational bottleneck for the nearest neighbor calculation with our smaller pool compared to theirs.

\textbf{Cluster Margin adapted to regression problem: Cluster Variance (ClusterVar)} \cite{citovsky2021batch}
To alleviate the robustness issue arising in larger batch AL scenarios, Cluster-Margin  \cite{citovsky2021batch} method is proposed to add necessary diversity to the uncertainty sampling. The original method used margin as its uncertainty metric as it was demonstrated on image classification tasks, and we adapted it into a variance metric in a regression setting. During Cluster Margin, Hierarchical Agglomerative Clustering (HAC) is run once on the unlabeled pool before the AL process and during each round a round-robin selection is carried out from the smallest cluster to the largest cluster, each time selecting the unlabeled sample with the highest uncertainty metric.


\section{Details of benchmark datasets used}

\begin{figure}[h!]
\vskip 0.2in
    \begin{center}
    \centerline{\includegraphics[width=0.6\columnwidth]{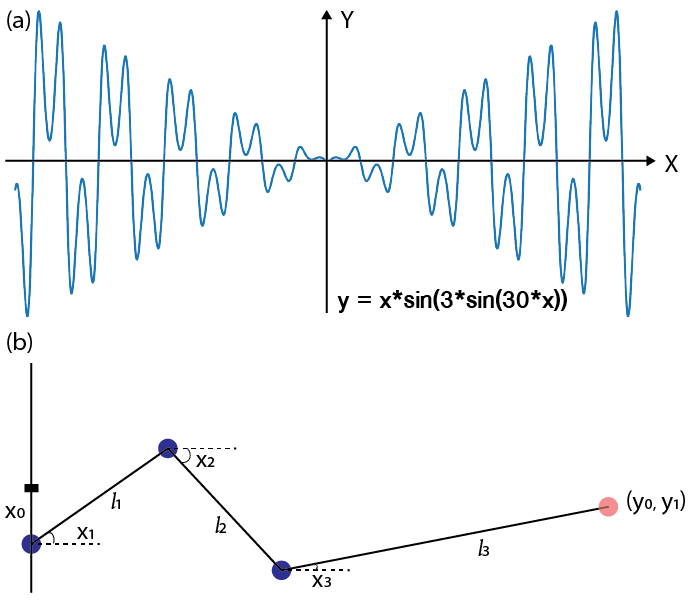}}
    \caption{Schematic illustration of sine wave (a) and robotic arm (b) datasets}
    \label{img:sup_data_toy}
    \end{center}
\vskip -0.2in
\end{figure}

\textbf{1D sine wave (Wave).} A noiseless 1-dimensional sinusoid with varying frequency over $x$, illustrated in \cref{img:sup_data_toy}.   
\begin{equation*}
    y = x * sin(a_1 * sin(a_2 * x)),
\end{equation*}
where $a_1 = 3$ and $a_2 = 30$ is chosen to make a relative complicated loss surface for the neural network to learn while also having a difference in sensitivity in the domain of x.

\textbf{2D robotic arm (Arm) \cite{ren2020benchmarking}} In this problem we aim to predict the 2D spatial location of the endpoint of a robotic arm based on its joint angles. Illustrated in \cref{img:sup_data_toy}.  The Oracle function is given by
\begin{equation*}
    y_0 = \sum_{i=1}^3 \cos(\dfrac{pi}{2}x_i)*l_i, 
    y_1 = x_0 + \sum_{i=1}^3 \sin(\dfrac{pi}{2}x_i)*l_i 
\end{equation*}
where $y$ is the position in the 2D plane, $x_0$ is the adjustable starting horizontal position, $x_{i=1,2,3}$ are the angles of the arm relative to horizontal reference and $l_{i=0,1,2} = [0.5, 0.5, 1]$ represents the i-th length of the robotic arm component. The dataset is available under the MIT license.

\begin{figure}[h!]
\vskip 0.2in
    \begin{center}
    \centerline{\includegraphics[width=0.6\columnwidth]{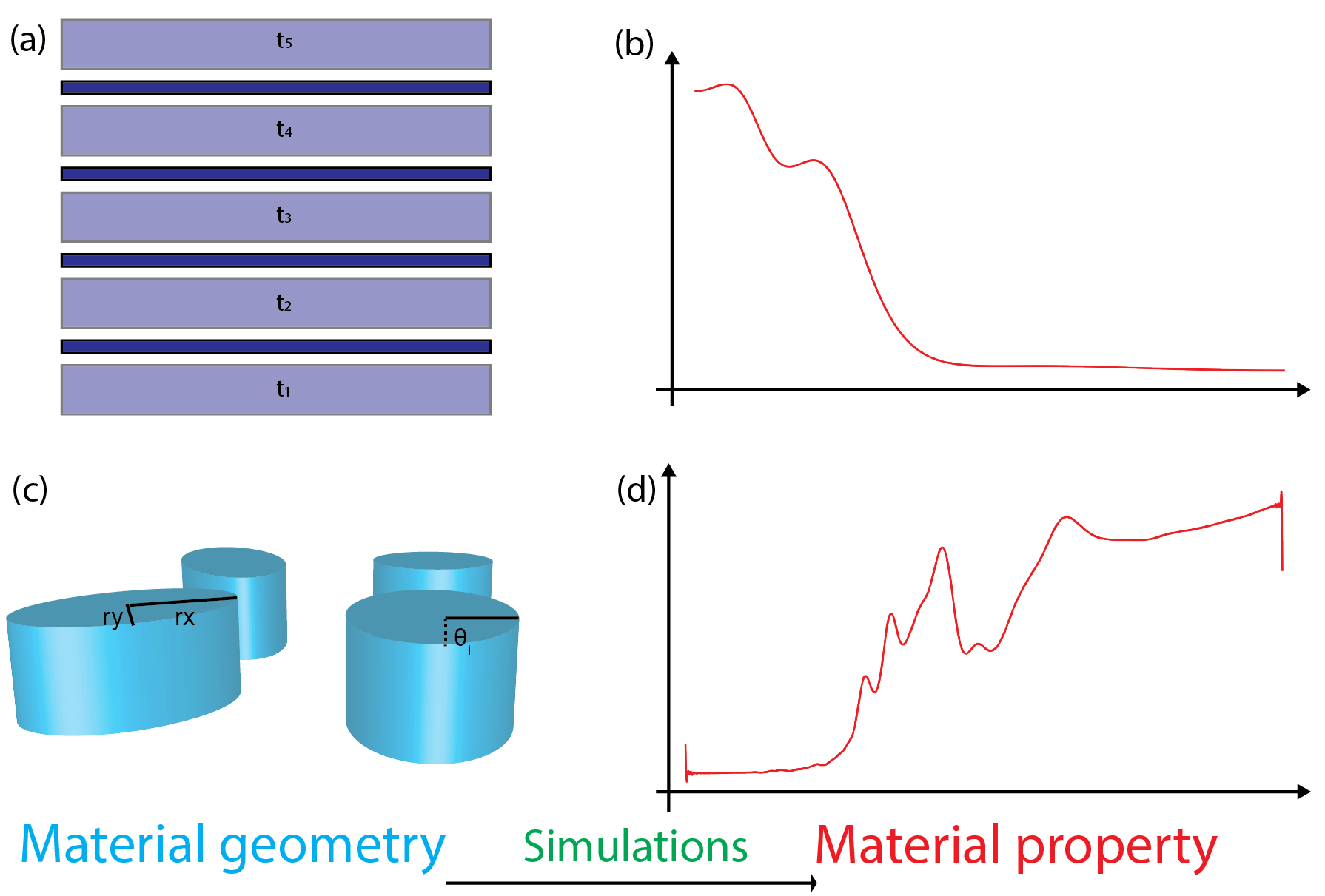}}
    \caption{(a, c) are schematic illustration of two material design datasets (Stack \& ADM). (b, d) are example spectra of their material property after simulations from their geometric parameterization (typically from Maxwell equation solvers that are slow and hence can benefit from active learning)}
    \label{img:sup_data_mm}
    \end{center}
\vskip -0.2in
\end{figure}

\textbf{Stacked material (Stack) \cite{Chen2019}.} In this problem, we aim to predict the reflection spectrum of a material, sampled at 201 wavelength points, based upon the thickness of each of the 5 layers of the material, illustrated in \cref{img:sup_data_mm}. It was also benchmarked in \cite{ren2022inverse}. An analytic Oracle function is available based upon physics\cite{Chen2019}.

\textbf{Artificial Dielectric Material (ADM) \cite{deng2021neural}} This problem takes the geometric structure of a material as input, and the reflection spectrum of the material, as a function of frequency, illustrated in \cref{img:sup_data_mm}. It was also benchmarked in \cite{deng2021benchmarking}. This dataset -released under CC BY 4.0 License - consists of input space of 3D geometric shape parameterized into 14 dimension space and the output is the spectral response of the material. The oracle function is a DNN \cite{deng2021benchmarking}. 

\begin{figure}[h!]
\vskip 0.2in
    \begin{center}
    \centerline{\includegraphics[width=0.6\columnwidth]{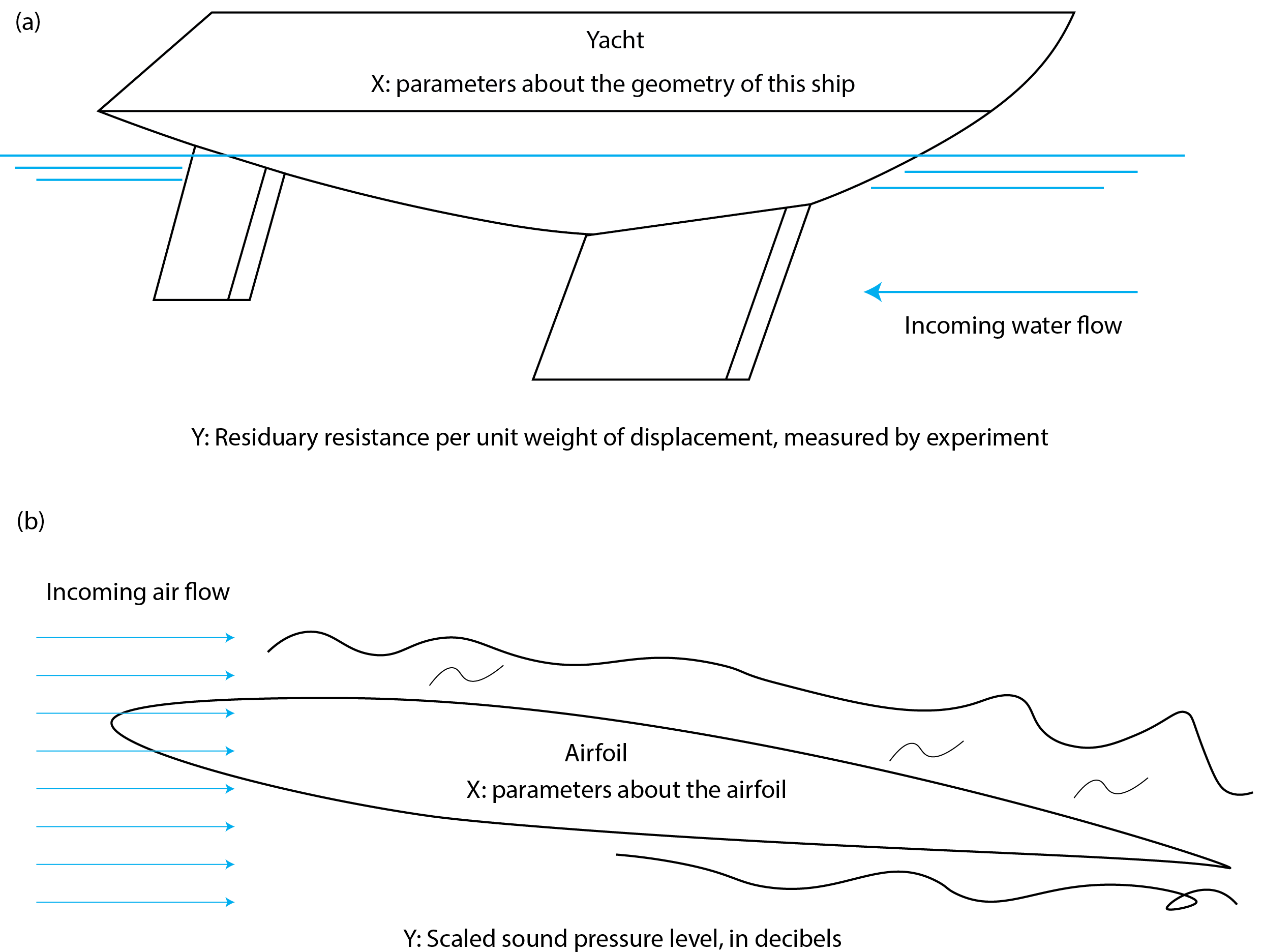}}
    \caption{Schematic illustration of Airfoil and Hydro experiments. Reproduced from the original source of experiment reports from NASA and Delft University of technology. (a) The Hydro experiment with an actual yacht being built and resistance was measured in a water flow experiment as the regression target y. (b) Airfoil experiment where input is the parameters of the airfoil and the sound pressure level is measured as target $y$ of the regression task.}
    \label{img:sup_UCI}
    \end{center}
\vskip -0.2in
\end{figure}

\textbf{NASA Airfoil (Foil) \cite{Dua:2019}} NASA dataset published on \url{https://archive.ics.uci.edu/dataset/291/airfoil+self+noise} UCI ML repository under CC BY 4.0 License \cite{Dua:2019} obtained from a series of aerodynamic and acoustic tests of 2D/3D airfoil blade sections conducted in an anechoic wind tunnel, illustrated in \cref{img:sup_UCI}. The input is the physical properties of the airfoil, like the angle of attack and chord length and the regression target is the sound pressure in decibels. We use a well-fitted random forest fit to the original dataset as our oracle function following prior work\cite{trabucco2022design}. The fitted random forest architecture and its weights are also shared in our code repo to ensure future work makes full use of such benchmark datasets as we did.

\textbf{Hydrodynamics (Hydro) \cite{Dua:2019}} Experiment conducted by the Technical University of Delft, illustrated in \cref{img:sup_UCI}, (hosted on \url{https://archive.ics.uci.edu/ml/datasets/Yacht+Hydrodynamics} UCI ML repository under CC BY 4.0 License \cite{Dua:2019}), this dataset contains basic hull dimensions and boat velocity and their corresponding residuary resistance. Input is 6 dimensions and output is the 1 dimension. We use a well-fitted random forest fit to the original dataset as our oracle function. The fitted random forest architecture and its weights are also shared in our code repository to ensure future work makes full use of such benchmark dataset as we did.

\begin{figure}[h!]
\vskip 0.2in
    \begin{center}
    \centerline{\includegraphics[width=0.6\columnwidth]{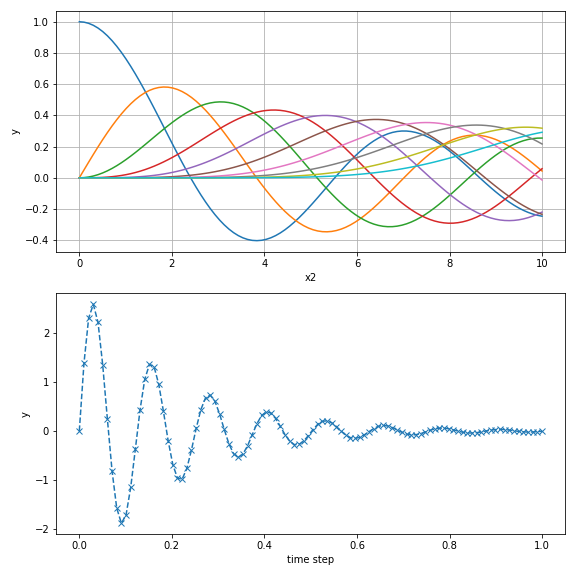}}
    \caption{Schematic illustration of Bessel function solution and the damping oscillator solutions.}
    \label{img:sup_ode}
    \end{center}
\vskip -0.2in
\end{figure}

\textbf{Bessel equation}
The solution to the below single dimension second-order differential equation:
\begin{equation*}
    x^2\dfrac{d^2y}{dx^2} + x\dfrac{dy}{dx} + (x^2 - \alpha^2)y=0
\end{equation*}
where input is $\alpha$ and $x$ position given. $\alpha$ is limited to non-negative integers smaller than 10 and $x \in [0, 10]$. The solution examples can be visualized in \cref{img:sup_ode}. Our choice of $\alpha$ values makes the Bessel functions cylinder harmonics and they frequently appear in solutions to Laplace's equation (in cylindrical systems). The implementation we used is the python package 'scipy.special.jv(v,z)' \cite{2020SciPy-NMeth}.

\textbf{Damping oscillator equation}
The solution to the below ordinary second-order differential equation:
\begin{equation*}
    m\dfrac{dx^2}{d^2t} + b\dfrac{dx}{dt} + \dfrac{mg}{l}x = 0
\end{equation*}
where m is the mass of the oscillator, b is the air drag, g is the gravity coefficient, l is the length of the oscillator's string and it has analytical solution of form 
\begin{equation*}
    x = a e^{-bt} cos(\alpha - \psi)
\end{equation*}
where a is the amplitude, b is the damping coefficient, $\alpha$ is the frequency and $\psi$ is the phase shift.  We assume $\psi$ to be 0 and let a,b,$\alpha$ be the input parameters. The output, unlike our previous ODE dataset, is taken as the first 100 time step trajectory of the oscillator, making it a high dimensional manifold (nominal dimension of 100 with true dimension of 3). The trajectory is illustrated in \cref{img:sup_ode}. We implement the above solution by basic python math operations.

\section{List of pool ratio used in existing literature}
 17000 \cite{mccallumzy1998employing}, 20 to 2000 \cite{kee2018query}, 300 to 375\cite{santos2020modeling}, 11-20 \cite{roy2018deep}, 1000 \cite{burbidge2007active}, and 1 to 11 \cite{tan2019batch}. 

\section{Details of models training and architecture}
In the below \cref{tbl:model_architectures}, we present the model architecture for each of our benchmarked datasets. Unless otherwise noted, all of them are fully connected neural networks.
\begin{table}[H]
    \caption{Regression model, $\hat{f}$ architecture details for each problem. *: for ADM, there are 3 layers of convolutions after the linear layer) }
    \label{tbl:model_architectures}
    \vskip 0.15in
    \begin{center}
        \begin{small}
        \begin{sc}
            \begin{tabular}{lcccccccc}
            \toprule
            Feat & Sine & Robo & Stack  & ADM & Foil & Hydr & BESS & DAMP\\
            \midrule
            NODE & 20 & 500 & 700 & 1500 & 200 & 50 & 50 & 500\\
            LAYER & 9 & 4 & 9 & $4^{*}$ & 4 & 6 & 6 & 6\\
            \bottomrule
            \end{tabular}
        \end{sc}
        \end{small}
    \end{center}
    \vskip -0.1in
\end{table}

We implemented our models in PyTorch \cite{NEURIPS2019_9015}. Beyond the above architectural differences, the rest of the model training settings are the same across the models: Starting labeled set of size 80, in each step DAL finds 40 points to be labeled for 50 active learning steps. Each regression model is an ensemble network of 10 models of size illustrated in \cref{tbl:model_architectures} except the ADM dataset (5 instead of 10 due to RAM issue). The test dataset is kept at 4000 points uniformly sampled across the $x$-space and they are fixed the same across all experiments for the same dataset. No bootstrapping is used to train the ensemble network and the only source of difference between networks in the ensemble (committee) is the random initialization of weights. 

The batch size is set to be 5000 (larger than the largest training set) so that the incomplete last batch would not affect the training result (as we sample more and more data, we can't throw away the last incomplete batch but having largely incomplete batch de-stabilizes training and introduce noise into the experiment. Adam optimizer is used with 500 training epochs and the model always retrains from scratch. (We observe that the training loss is much higher if we inherit from the last training episode and do not start from scratch, which is consistent with other literature \cite{beck2021effective}). The learning rate is generally 1e-3 (some datasets might differ), and the decay rate of 0.8 with the decay at the plateau training schedule. The regularization weight is usually 1e-4 (dataset-dependent as well). The hyper-parameters only change with respect to the dataset but never with respect to DAL used. 

The hyperparameters are tuned in the same way as the model architecture: Assume we have a relatively large dataset (2000 randomly sampling points) and tune our hyperparameter on this set. This raises another robustness problem of deep active learning, which is how to determine the model architecture before we have enough labels. This is currently out of the scope of this work as we focused on how different DALs behave with the assumption that the model architectures are chosen smartly and would be happy to investigate this issue in future work.

For the BALD method, we used a dropout rate of 0.5 as advised by previous work. As BALD requires a different architecture than other base methods (a dropout structure, that is capable of getting a good estimate even with 50\% of the neurons being dropped), the model architecture for the active learning is different in that it enlarges each layer by a constant factor that can make it the relatively same amount of total neurons like other DAL methods. Initially, the final trained version of the dropout model is used as the regression model to be evaluated. However, we found that an oversized dropout model hardly fits as well as our ensembled counterpart like other DAL methods. Therefore, to ensure the fairness of comparison, we trained another separate, ensembled regression model same as the other DALs and reported our performance on that.

For the LearningLoss method, we used the same hyper-parameter that we found in the cited work in the main text: relative weight of learning loss of 0.001 and a training schedule of 60\% of joint model training and the rest epoch we cut the gradient flow to the main model from the auxiliary model. For the design of the auxiliary model, we employed a concatenation of the output of the last three hidden layers of our network, each followed by a fully connected network of 10 neurons, before being directed to the final fully connected layer of the auxiliary network that produces a single loss estimate.

For the EMOC method, due to RAM limit and time constraint, we can not consider all the model parameters during the gradient calculation step (For time constraint, \cref{tbl:time_performance} gives a good reference of how much longer EMOC cost, even in this reduced form). Therefore, we implemented two approximations: (i) For the training set points where the current model gradients are evaluated, instead of taking the ever-growing set that is more and more biased towards the DAL selection, we fixed it to be the 80 original, uniformly sampled points. (ii) We limit the number of model parameters to evaluate the EMOC criteria to 50k. We believe taking the effect of 50 thousand parameters gives a good representation of the model's response (output change) for new points. We acknowledge that these approximations might be posing constraints to EMOC, however, these are practical, solid challenges for DAL practitioners as well and these are likely the compromise to be made during application.

\subsection{Computational resources}
Here we report the computational resources we used for this work: AMD CPU with 64 cores; NVIDIA 3090 GPU x4 (for each of the experiment we used a single GPU to train); 256GB RAM.

\section{Additional performance plots}
As the benchmark conducts a huge set of experiments that are hard to fit in the main text, here we present all the resulting figures for those who are interested to dig more takeaways.

\subsection{Time performance of the benchmarked DAL methods}
We also list the time performance of each DAL method, using the ROBO dataset as an example in the below  \cref{tbl:time_performance}. Note that this is only the sampling time cost, not including the model training time, which is usually significantly larger than the active learning sampling time at each step. The only DAL method that potentially has a time performance issue is the EMOC method, which requires the calculation of each of the gradients with respect to all parameters and therefore takes a much longer time than other DAL methods. However, as it is shown in the main text that it is not a robust method in our setting, there is no dilemma of performance/time tradeoff presented here.

\begin{table}[h]
    \caption{Time performance for average time spent during the sampling process for ROBO dataset per active learning step (40 points) in ms for pool ratio of 2. LL: LearningLoss }
    \label{tbl:time_performance}
    \begin{center}
        \begin{small}
        \begin{sc}
            \begin{tabular}{lcccccccccc}
            \toprule
            Dataset & Random & GSx & GSxy & GSy & BALD & EMOC  \\
            \midrule
            Time & 2.15 & 4.96 & 10.27 & 6.85 & 9.06 & \textbf{756.6}  \\
            \midrule
             Dataset & LL & QBC & QBCDiv & QBCDivDen & DACS &  ClusterVar  \\
            \midrule
             Time & 6.53 & 4.29 & 9.04 & 10.38 & 21.70 & 4.31 \\
            \bottomrule
            \end{tabular}
        \end{sc}
        \end{small}
    \end{center}
\end{table}

\subsection{Combined plot with $nAUC_{MSE}$ and nDiv}

\begin{figure}[h!]
\vskip -0.2in
    \begin{center}
    \centerline{\includegraphics[width=0.9\columnwidth]{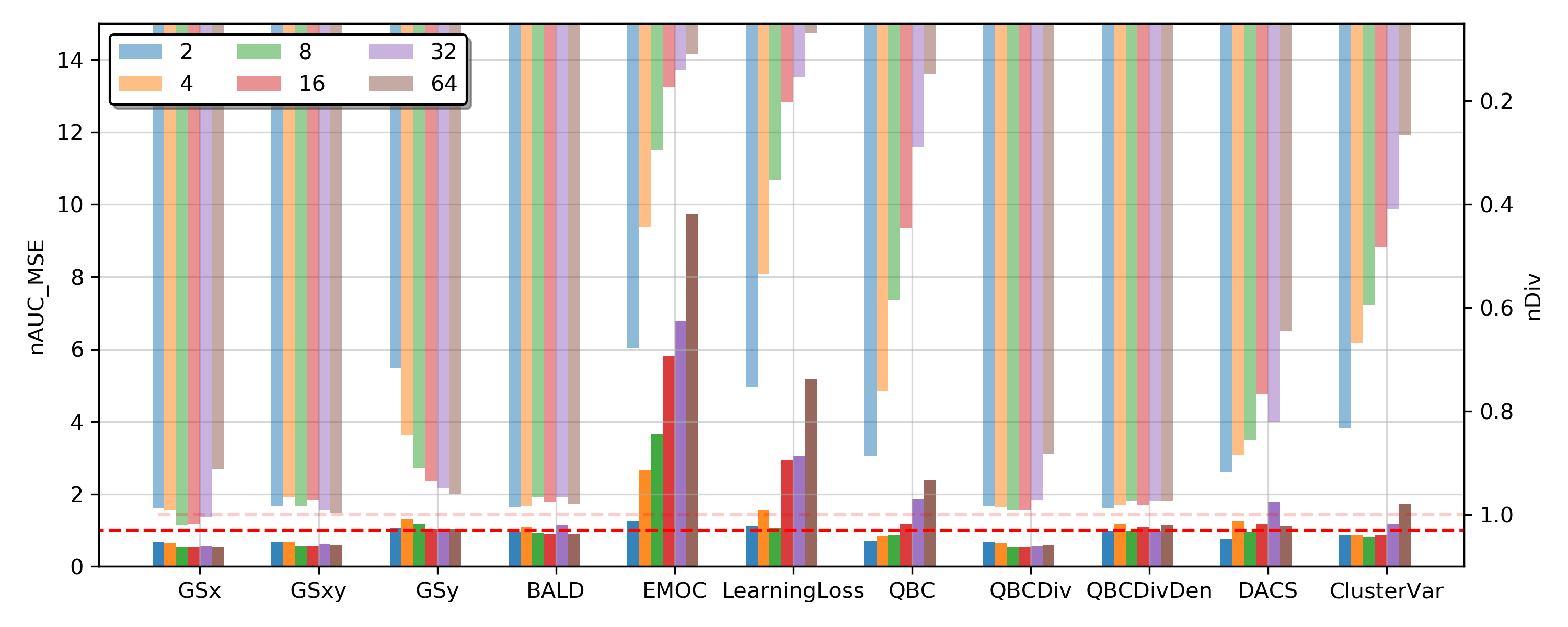}}
    \caption{$nAUC_{MSE}$ and nDiv plot for SINE}
    \end{center}
\vskip -0.2in
\end{figure}

\begin{figure}[h!]
\vskip -0.2in
    \begin{center}
    \centerline{\includegraphics[width=0.9\columnwidth]{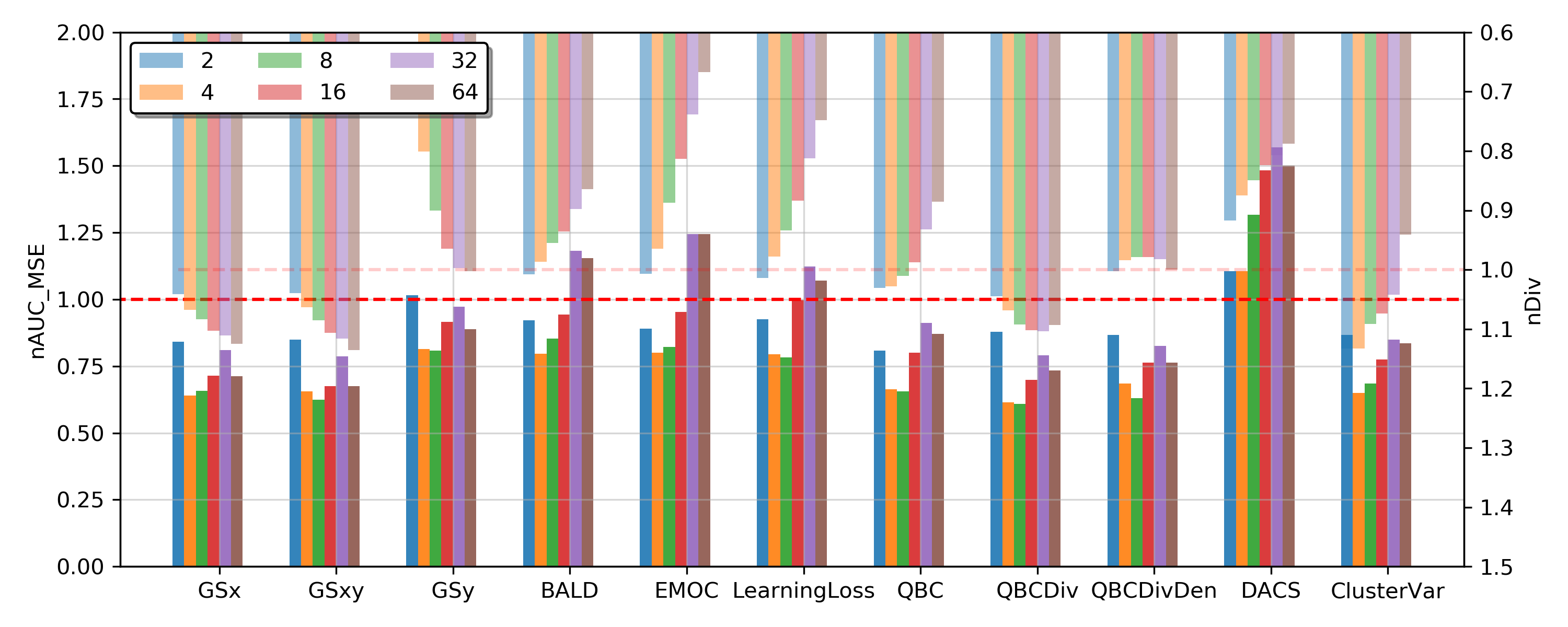}}
    \caption{$nAUC_{MSE}$ and nDiv plot for STACK}
    \end{center}
\vskip -0.2in
\end{figure}

\begin{figure}[h!]
\vskip -0.2in
    \begin{center}
    \centerline{\includegraphics[width=0.9\columnwidth]{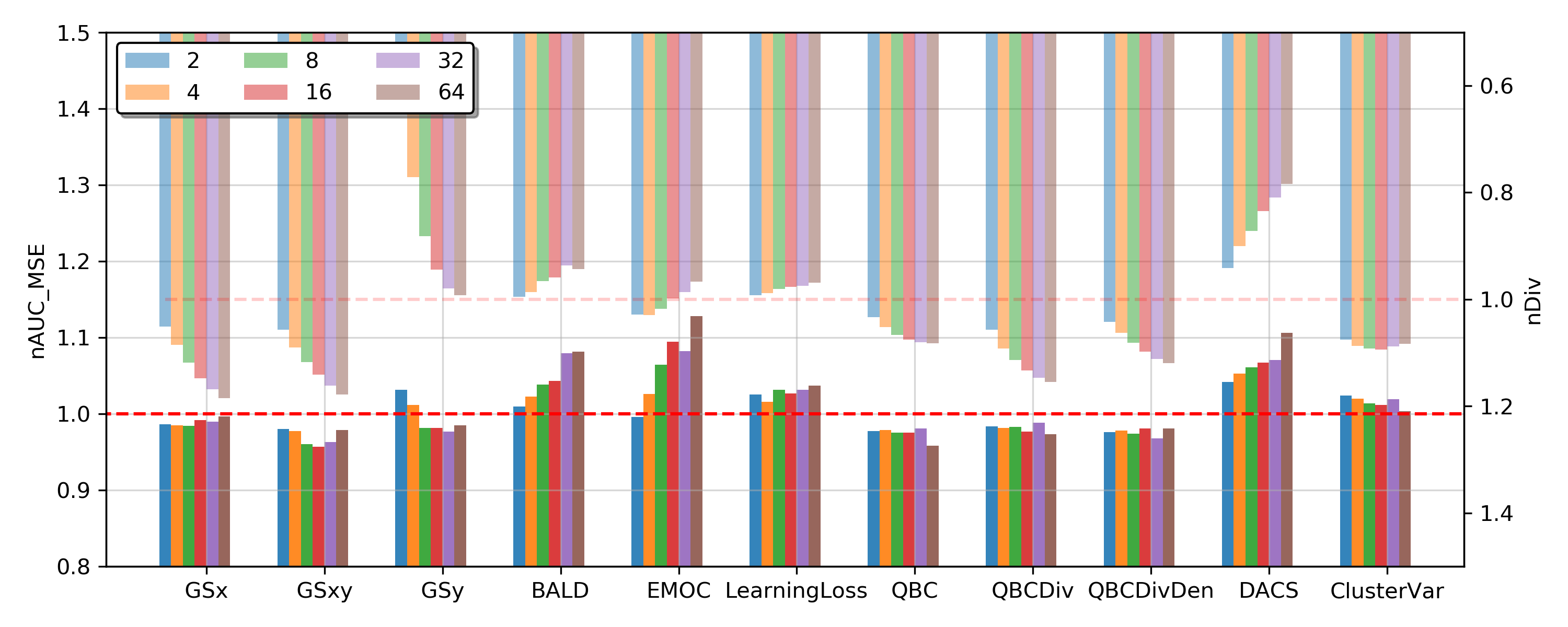}}
    \caption{$nAUC_{MSE}$ and nDiv plot for ADM}
    \end{center}
\vskip -0.2in
\end{figure}

\begin{figure}[h!]
\vskip -0.2in
    \begin{center}
    \centerline{\includegraphics[width=0.9\columnwidth]{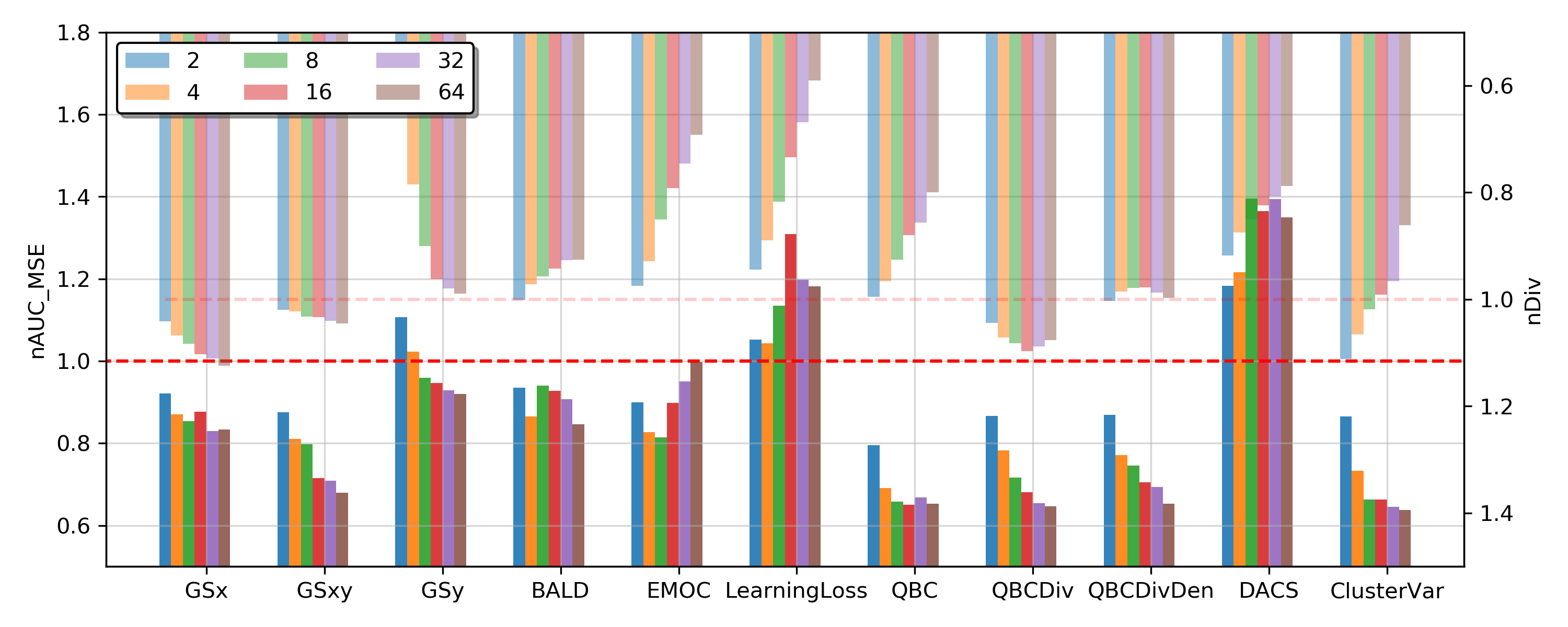}}
    \caption{$nAUC_{MSE}$ and nDiv plot for FOIL}
    \end{center}
\vskip -0.2in
\end{figure}

\begin{figure}[h!]
\vskip -0.2in
    \begin{center}
    \centerline{\includegraphics[width=0.9\columnwidth]{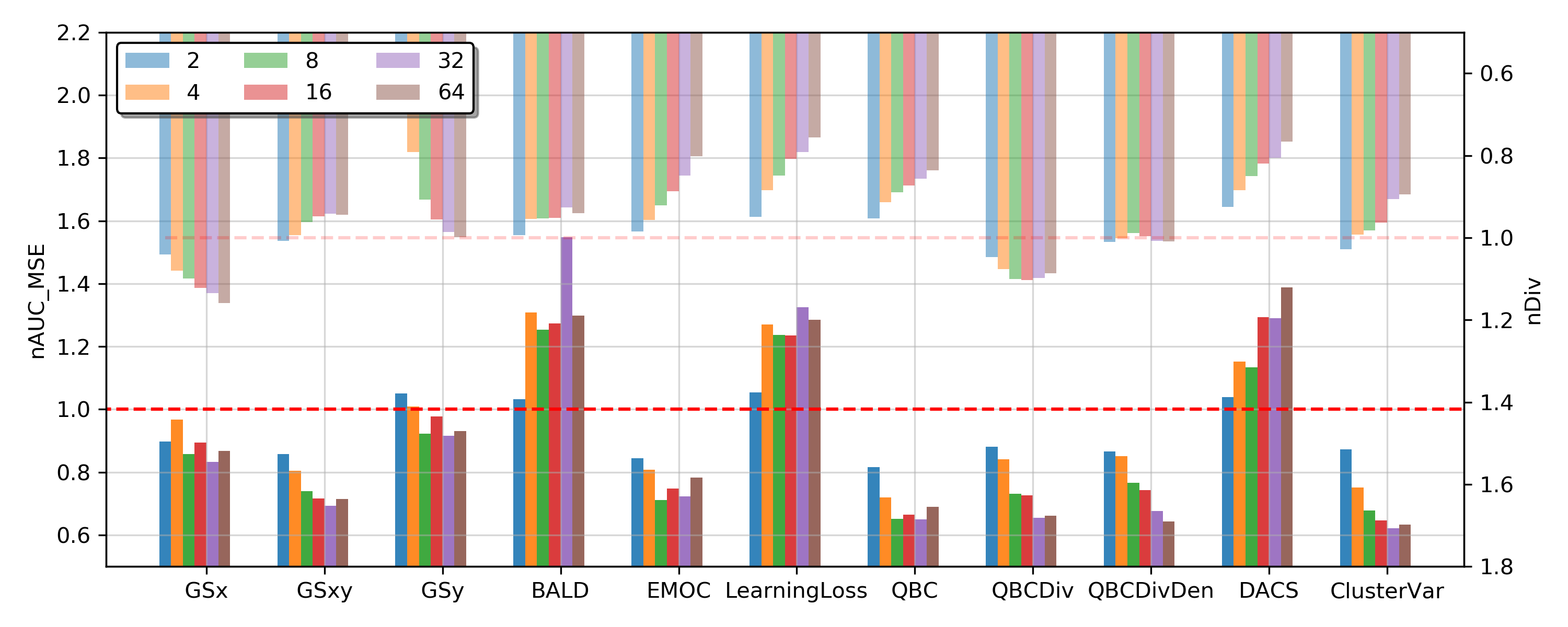}}
    \caption{$nAUC_{MSE}$ and nDiv plot for HYDR}
    \end{center}
\vskip -0.2in
\end{figure}

\begin{figure}[h!]
\vskip -0.2in
    \begin{center}
    \centerline{\includegraphics[width=0.9\columnwidth]{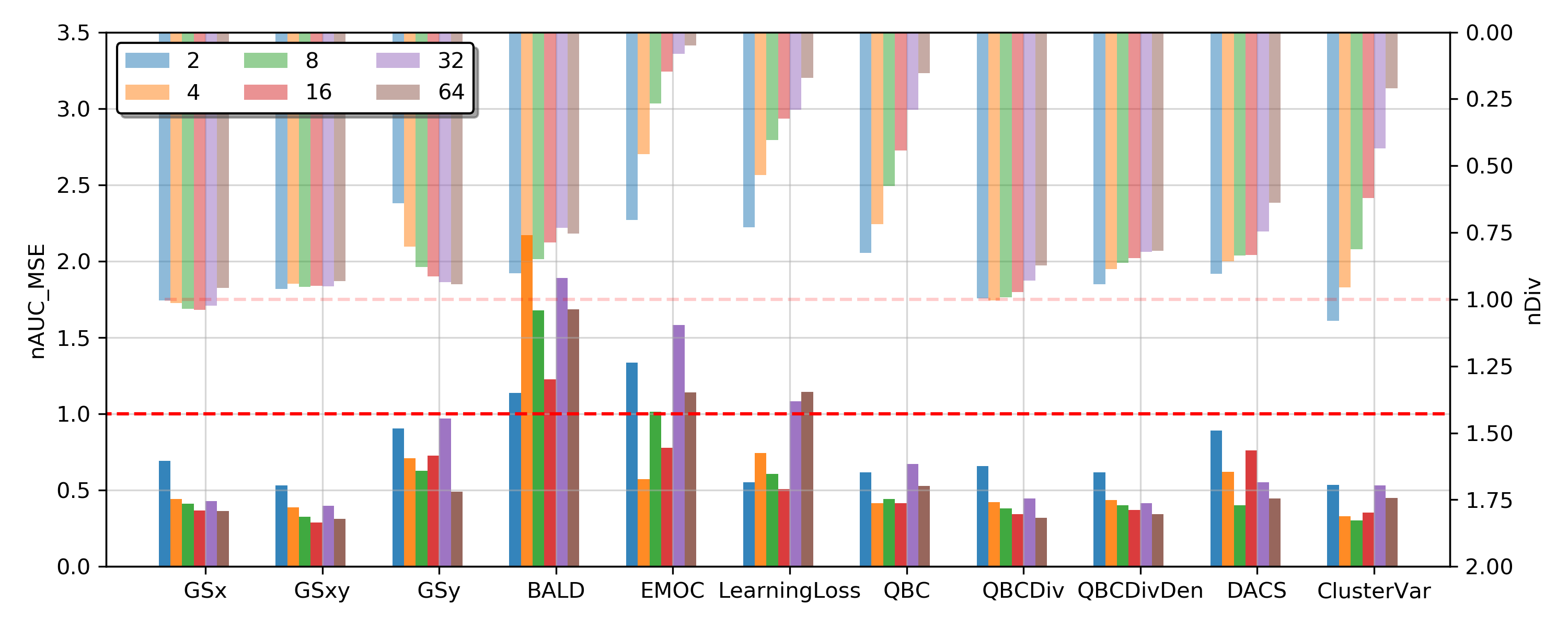}}
    \caption{$nAUC_{MSE}$ and nDiv plot for BESS}
    \end{center}
\vskip -0.2in
\end{figure}

\begin{figure}[h!]
\vskip -0.2in
    \begin{center}
    \centerline{\includegraphics[width=0.9\columnwidth]{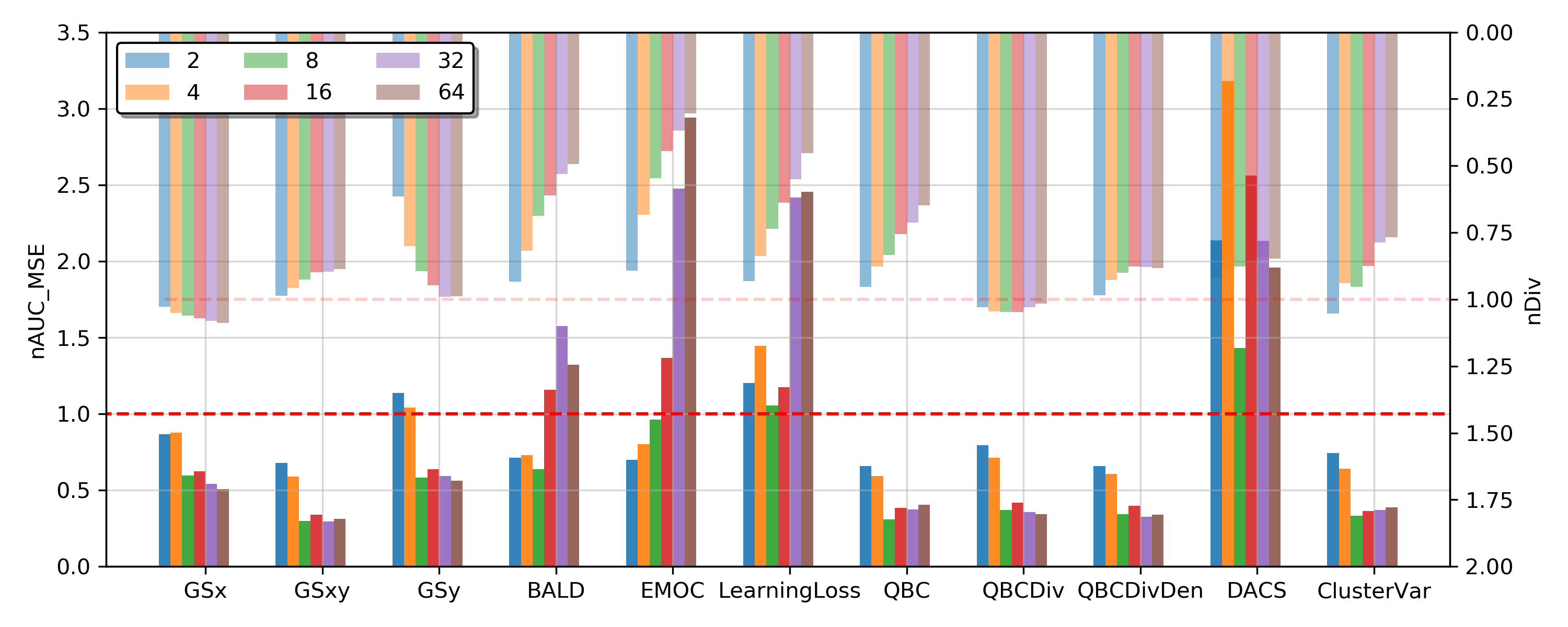}}
    \caption{$nAUC_{MSE}$ and nDiv plot for DAMP}
    \end{center}
\vskip -0.2in
\end{figure}

\subsection{MSE vs active learning step plot}
We also present the traditional plot of the MSE vs active learning step for reference. For each of the plots below, the MSE are smoothed with a smoothing parameter of 0.5 using the tensorboard smoothing visualizing function \cite{tensorflow2015-whitepaper}. The $x$ labels are from 0 - 49, where 0 measures the end of the first active learning step.

\begin{figure}[h!]
\vskip -0.2in
    \begin{center}
    \centerline{\includegraphics[width=\columnwidth]{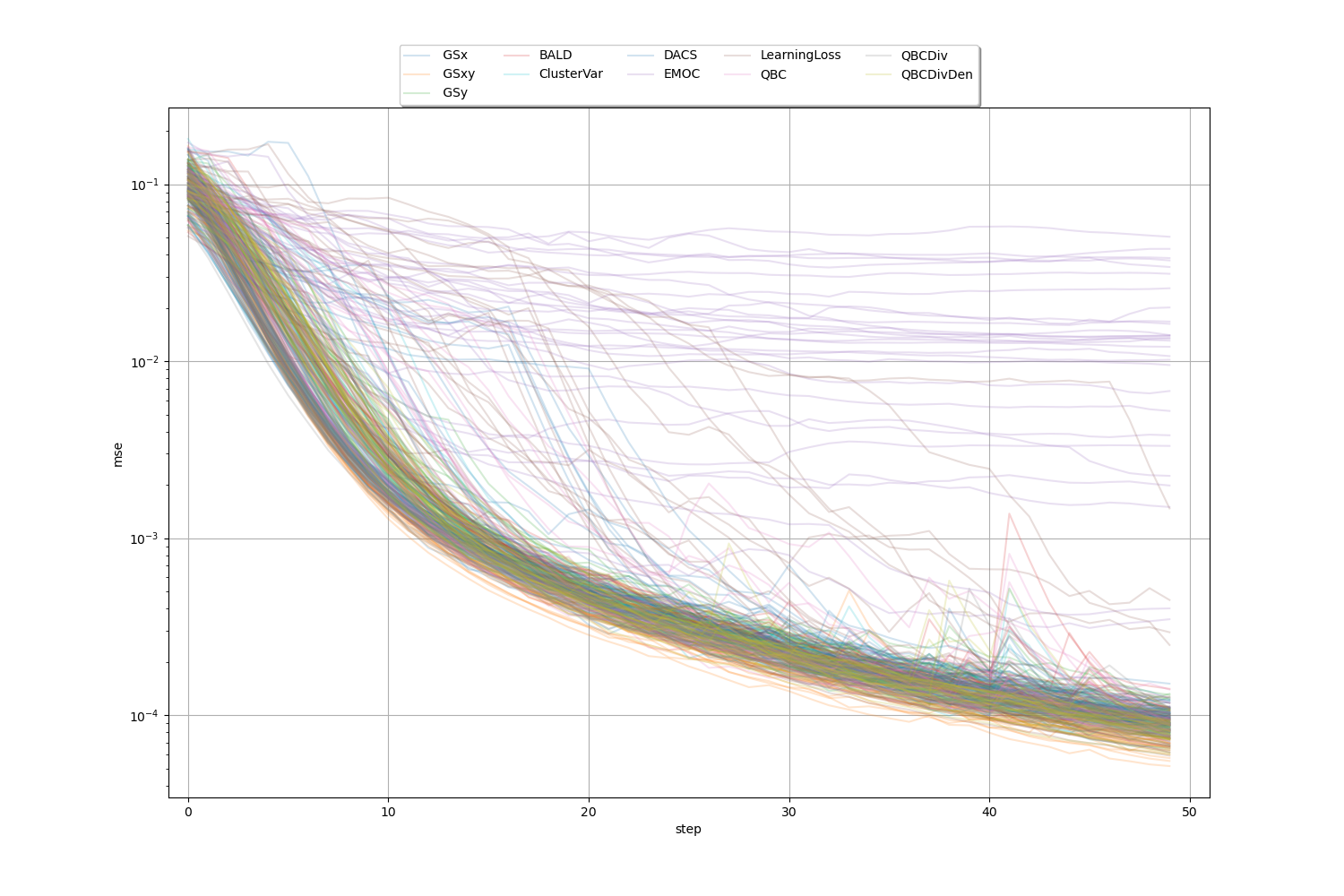}}
    \vskip -0.2in
    \caption{MSE plot for SINE}
    \end{center}
\vskip -0.2in
\end{figure}

\begin{figure}[h!]
\vskip -0.2in
    \begin{center}
    \centerline{\includegraphics[width=\columnwidth]{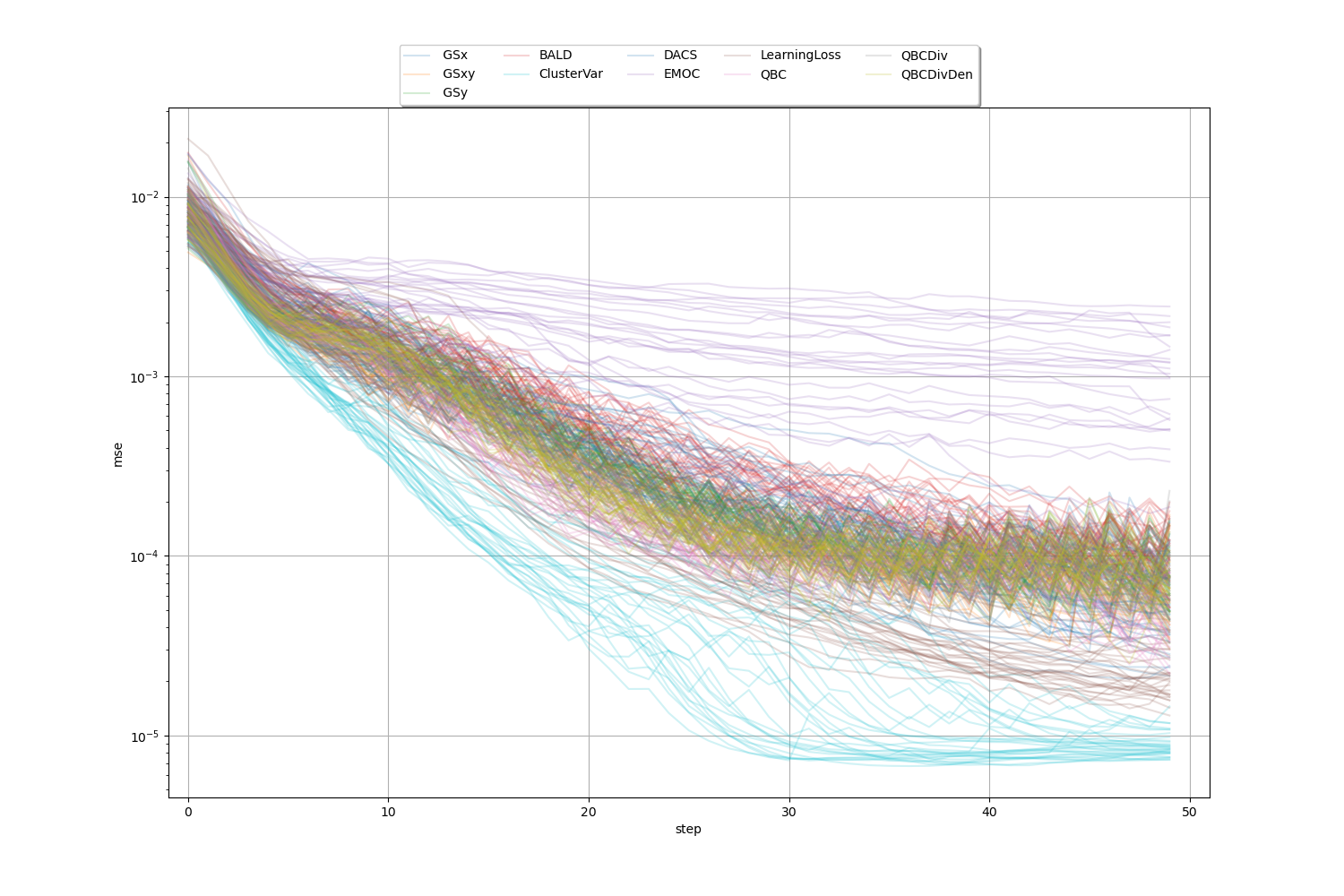}}
    \vskip -0.2in
    \caption{MSE plot for ROBO}
    \end{center}
\vskip -0.2in
\end{figure}

\begin{figure}[h!]
\vskip -0.2in
    \begin{center}
    \centerline{\includegraphics[width=\columnwidth]{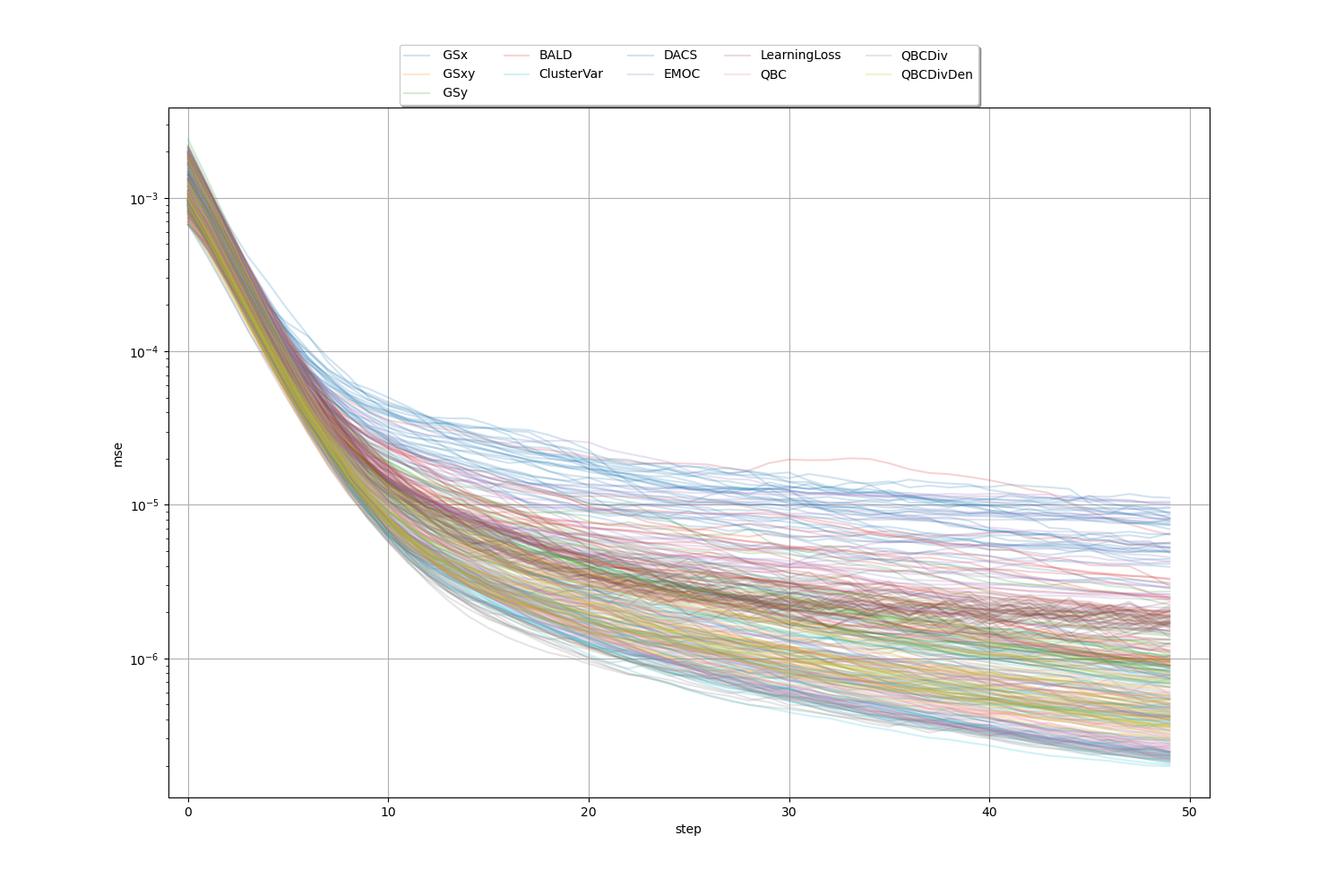}}
    \vskip -0.2in
    \caption{MSE plot for STACK}
    \end{center}
\vskip -0.2in
\end{figure}
\begin{figure}[h!]
\vskip -0.2in
    \begin{center}
    \centerline{\includegraphics[width=\columnwidth]{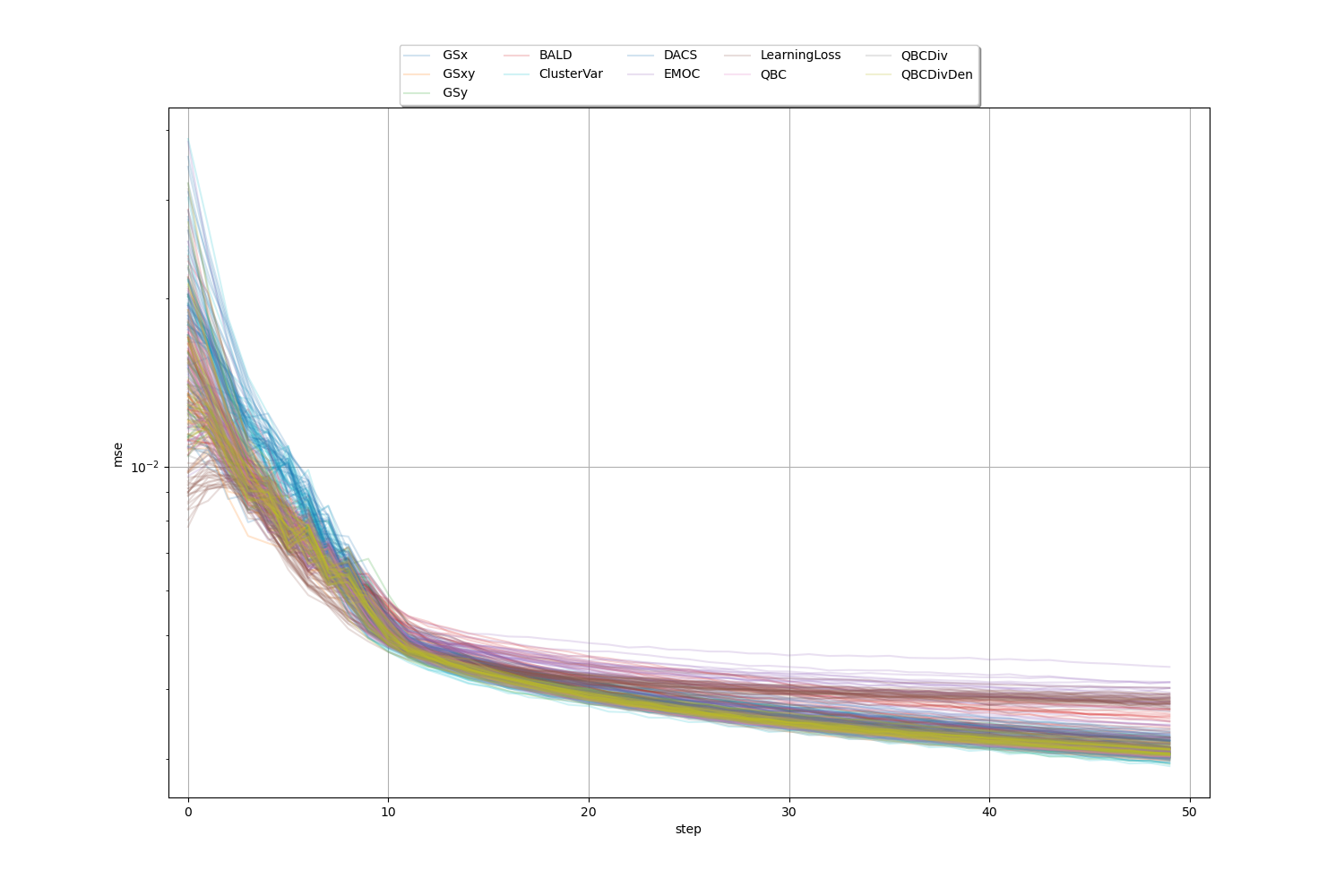}}
    \vskip -0.2in
    \caption{MSE plot for ADM}
    \end{center}
\vskip -0.2in
\end{figure}
\begin{figure}[h!]
\vskip -0.2in
    \begin{center}
    \centerline{\includegraphics[width=\columnwidth]{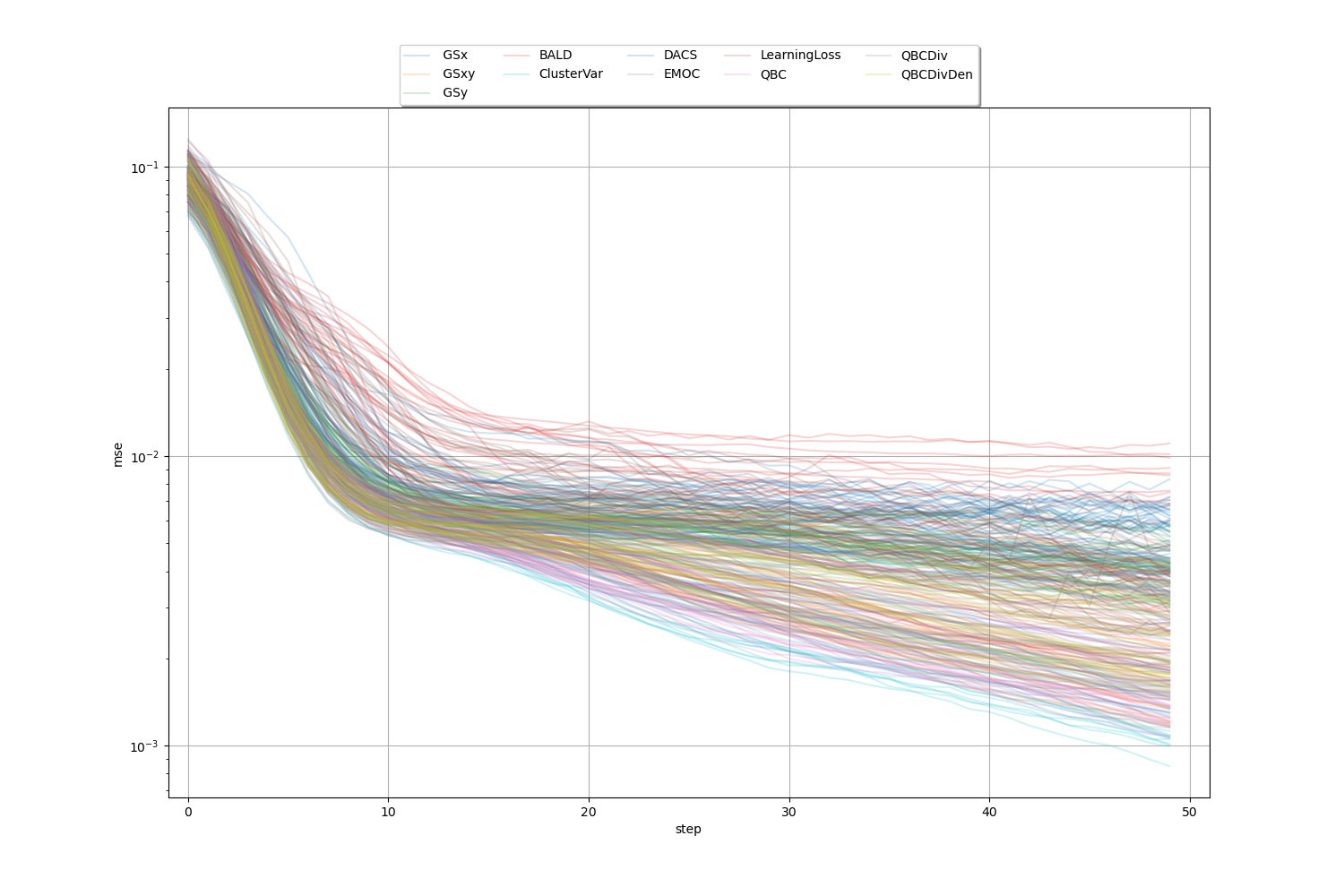}}
    \vskip -0.2in
    \caption{MSE plot for HYDR}
    \end{center}
\vskip -0.2in
\end{figure}
\begin{figure}[h!]
\vskip -0.2in
    \begin{center}
    \centerline{\includegraphics[width=\columnwidth]{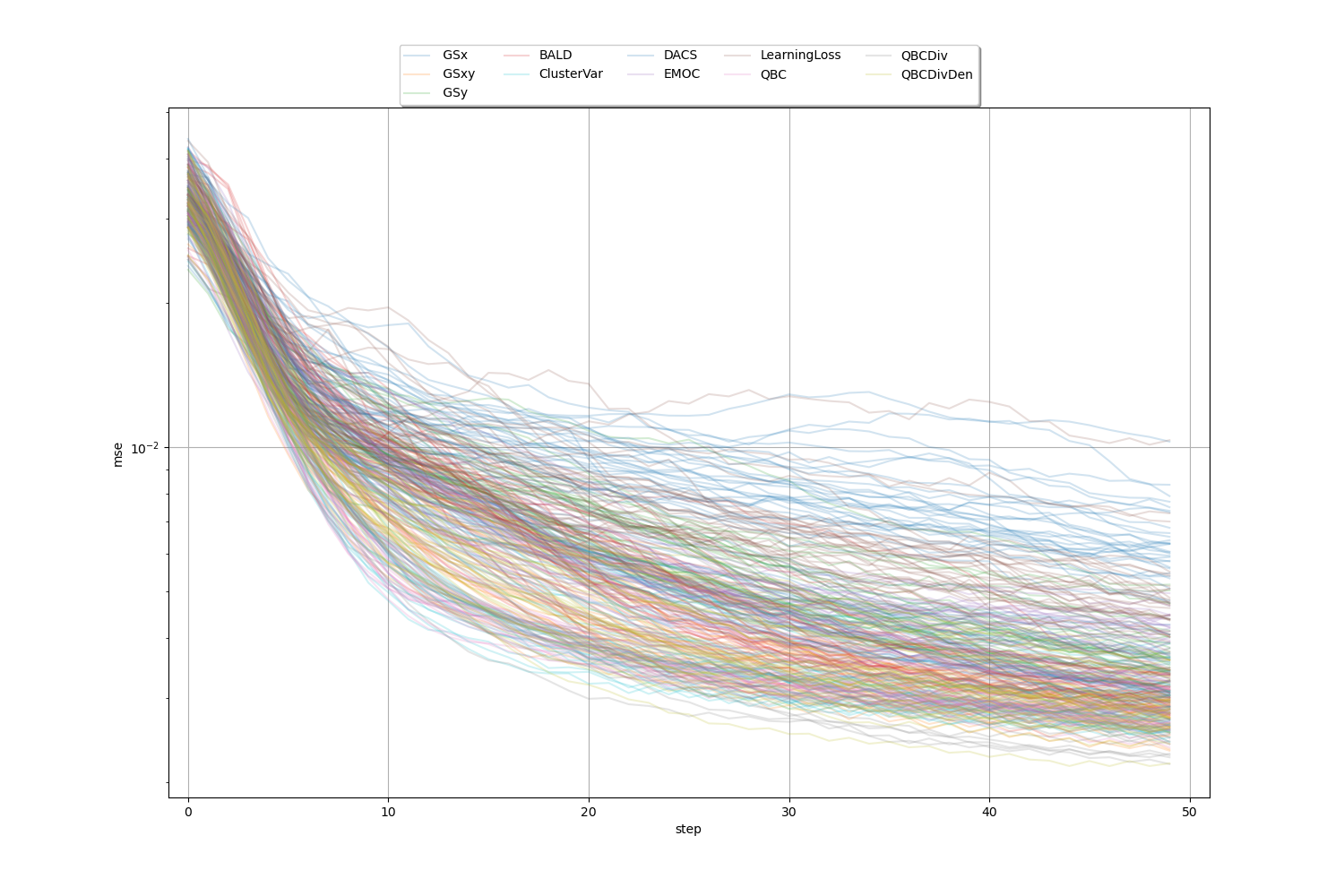}}
    \vskip -0.2in
    \caption{MSE plot for FOIL}
    \end{center}
\vskip -0.2in
\end{figure}
\begin{figure}[h!]
\vskip -0.2in
    \begin{center}
    \centerline{\includegraphics[width=\columnwidth]{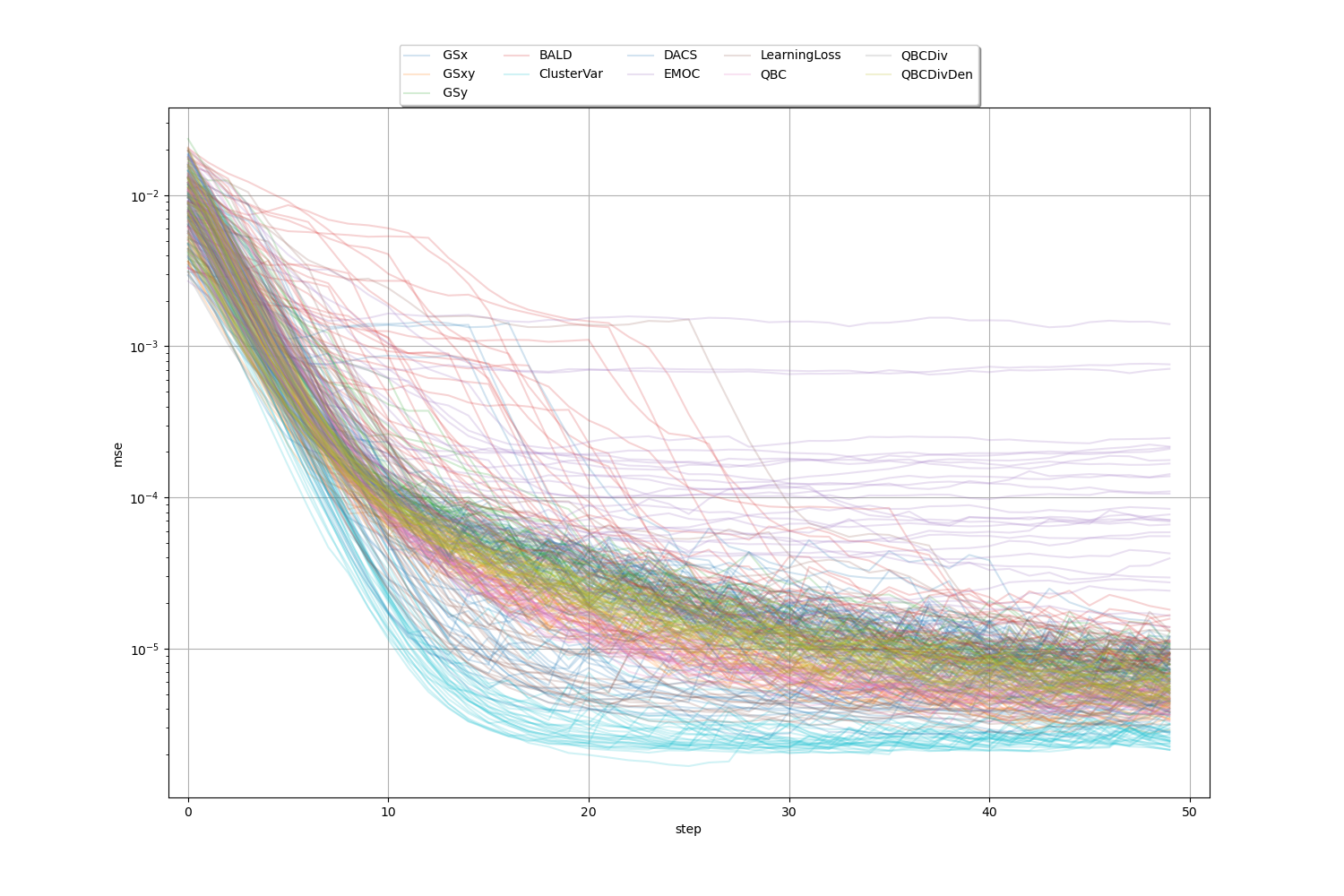}}
    \vskip -0.2in
    \caption{MSE plot for BESS}
    \end{center}
\vskip -0.2in
\end{figure}
\begin{figure}[h!]
\vskip -0.2in
    \begin{center}
    \centerline{\includegraphics[width=\columnwidth]{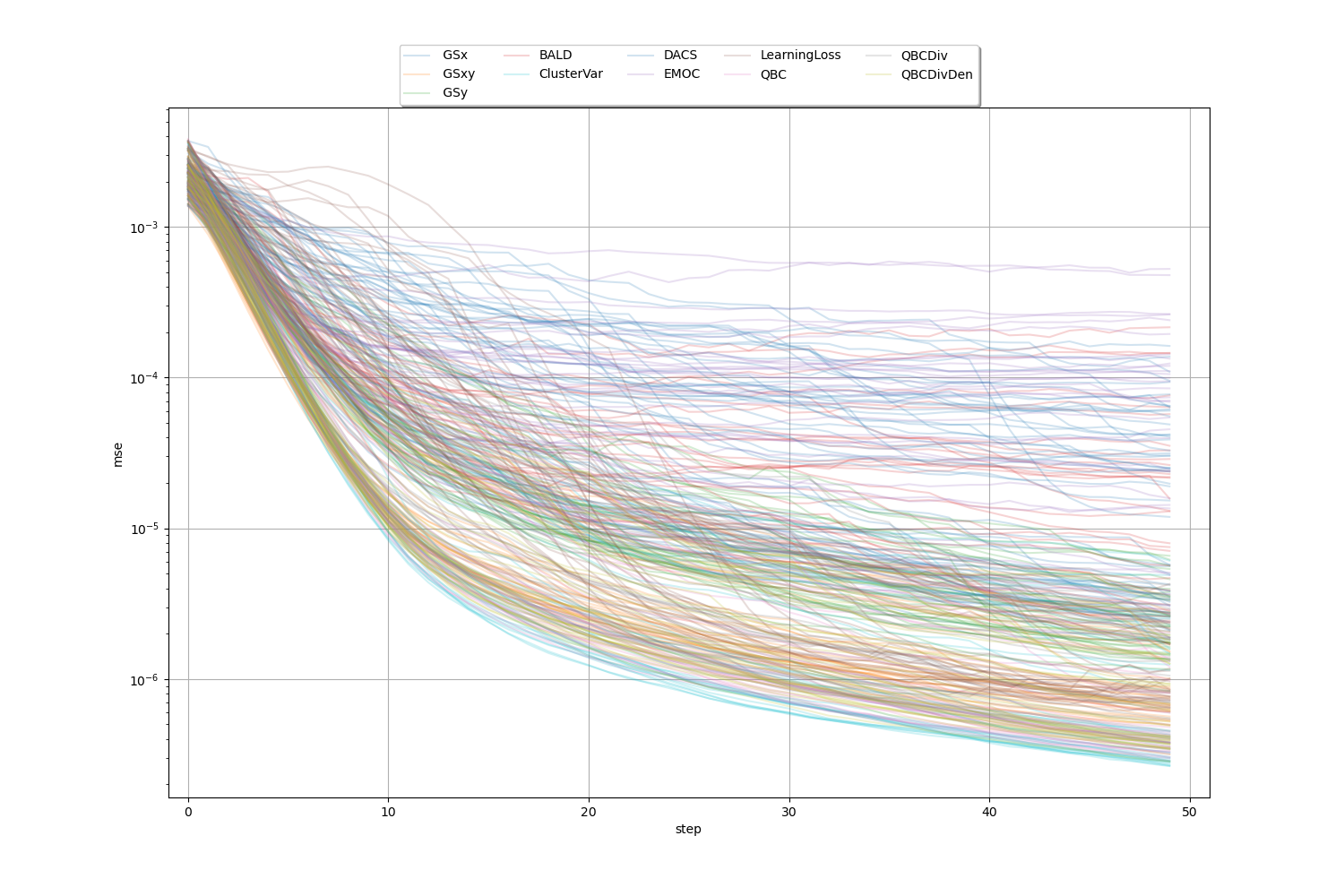}}
    \vskip -0.2in
    \caption{MSE plot for DAMP}
    \end{center}
\vskip -0.2in
\end{figure}

\end{document}